\documentclass[10pt,journal,compsoc]{IEEEtran}
\usepackage{graphicx}

\ifCLASSOPTIONcompsoc
  \usepackage[nocompress]{cite}
\else
  \usepackage{cite}
\fi
\usepackage{hyperref}       
\usepackage{url}            
\usepackage{epsfig}
\usepackage{graphicx}
\usepackage{booktabs}       
\usepackage{amsfonts}       
\usepackage{nicefrac}       
\usepackage{microtype}      
\usepackage{xcolor}         
\usepackage{amsmath}
\usepackage{amsfonts}
\usepackage{amssymb}
\usepackage{cleveref}
\usepackage{cancel}
\usepackage{amsthm}
\usepackage{wrapfig}
\usepackage{multirow}
\usepackage{subfig}
\usepackage{enumerate}
\usepackage{amsthm}
\usepackage{float}
\usepackage{cancel}
\usepackage[ruled,vlined]{algorithm2e}
\newtheorem{prop}{Proposition}
\newtheorem{duplicate}{Proposition}
\newtheorem{lemma}{Lemma}
\newtheorem{dup}{Lemma}
\usepackage{enumitem}
\usepackage{listings}
\setlist[itemize]{leftmargin=*}
\ifCLASSINFOpdf
\else
\fi
\def\mA{{\mathbf{A}}}
\def\mB{{\mathbf{B}}}
\def\mC{{\mathbf{C}}}
\def\mD{{\mathbf{D}}}

\def\mH{{\mathbf{H}}}
\def\mI{{\mathbf{I}}}

\def\mK{{\mathbf{K}}}

\def\mM{{\mathbf{M}}}

\def\mP{{\mathbf{P}}}
\def\mQ{{\mathbf{Q}}}
\def\mR{{\mathbf{R}}}
\def\mS{{\mathbf{S}}}
\def\mT{{\mathbf{T}}}
\def\mU{{\mathbf{U}}}
\def\mV{{\mathbf{V}}}

\def\mX{{\mathbf{X}}}
\def\mY{{\mathbf{Y}}}
\def\mZ{{\mathbf{Z}}}

\def\mLambda{{\mathbf{\Lambda}}}

\begin{document}
%
\title{Fast Differentiable Matrix Square Root and Inverse Square Root}
%
%
%
%
\author{Yue~Song,~\IEEEmembership{Member,~IEEE,}
        Nicu~Sebe,~\IEEEmembership{Senior Member,~IEEE,}
        Wei~Wang,~\IEEEmembership{Member,~IEEE}
\IEEEcompsocitemizethanks{\IEEEcompsocthanksitem Yue Song, Nicu Sebe, and Wei Wang are with the Department
of Information Engineering and Computer Science, University of Trento, Trento 38123,
Italy.\protect\\
E-mail: \{yue.song, nicu.sebe, wei.wang\}@unitn.it}
\thanks{Manuscript received April 19, 2005; revised August 26, 2015.}}

%
%

\markboth{IEEE TRANSACTIONS ON PATTERN ANALYSIS AND MACHINE INTELLIGENCE}%
{Shell \MakeLowercase{\textit{et al.}}: Bare Demo of IEEEtran.cls for Computer Society Journals}
%



\IEEEtitleabstractindextext{%
\begin{abstract}
Computing the matrix square root and its inverse in a differentiable manner is important in a variety of computer vision tasks. Previous methods either adopt the Singular Value Decomposition (SVD) to explicitly factorize the matrix or use the Newton-Schulz iteration (NS iteration) to derive the approximate solution. However, both methods are not computationally efficient enough in either the forward pass or the backward pass. In this paper, we propose two more efficient variants to compute the differentiable matrix square root and the inverse square root. For the forward propagation, one method is to use Matrix Taylor Polynomial (MTP), and the other method is to use Matrix Pad\'e Approximants (MPA). The backward gradient is computed by iteratively solving the continuous-time Lyapunov equation using the matrix sign function. A series of numerical tests show that both methods yield considerable speed-up compared with the SVD or the NS iteration. Moreover, we validate the effectiveness of our methods in several real-world applications, including de-correlated batch normalization, second-order vision transformer, global covariance pooling for large-scale and fine-grained recognition, attentive covariance pooling for video recognition, and neural style transfer. The experiments demonstrate that our methods can also achieve competitive and even slightly better performances. Code is available at \href{https://github.com/KingJamesSong/FastDifferentiableMatSqrt}{https://github.com/KingJamesSong/FastDifferentiableMatSqrt}.


\end{abstract}


}

\maketitle

\IEEEdisplaynontitleabstractindextext

%
\IEEEpeerreviewmaketitle


%

\IEEEraisesectionheading{\section{Introduction}\label{sec:introduction}}

\begin{figure}[tbp]
    \centering
    \includegraphics[width=0.8\linewidth]{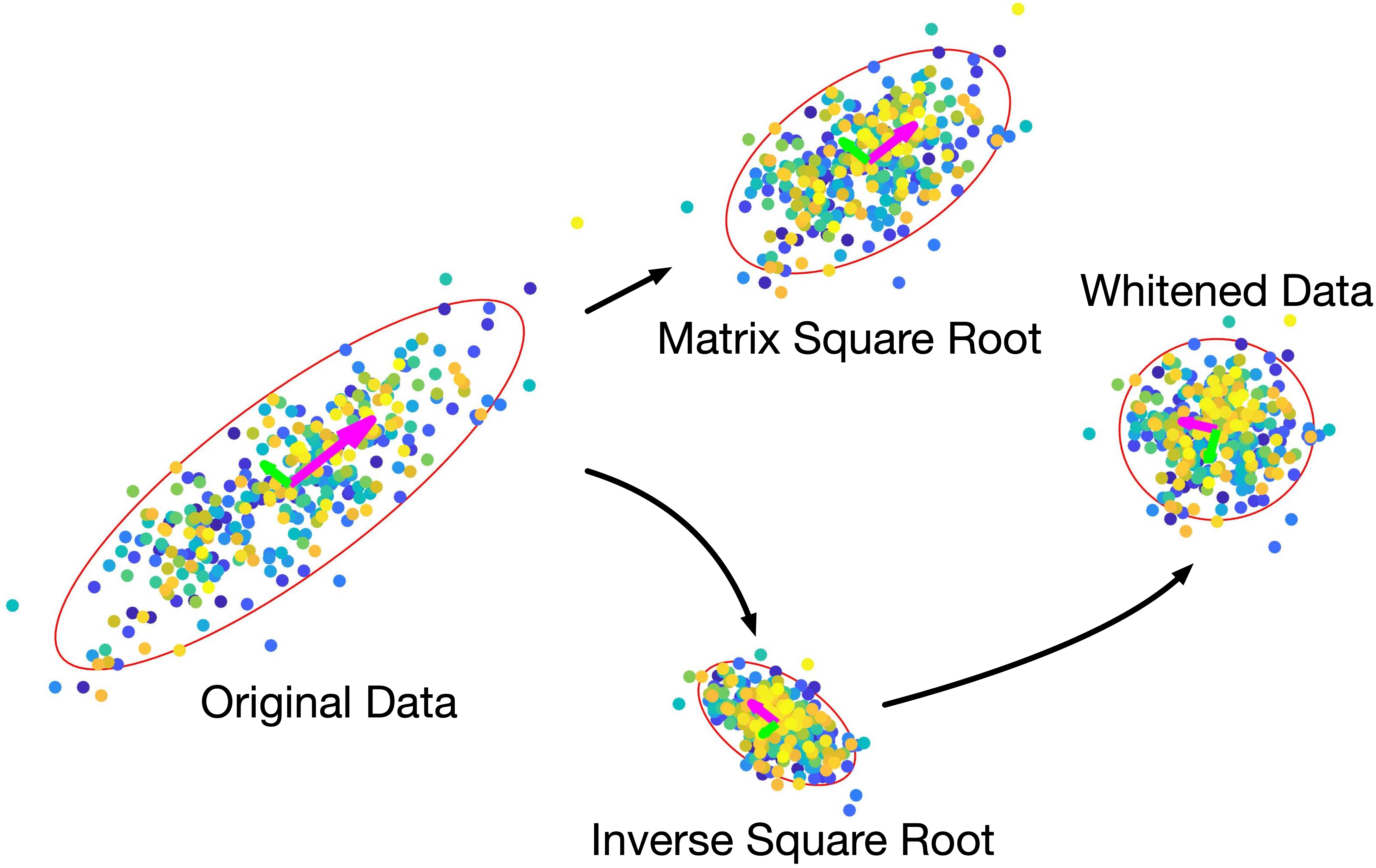}
    \caption{Exemplary visualization of the matrix square root and its inverse. Given the original data $\mX{\in}\mathbb{R}^{2{\times}n}$, the matrix square root performs an effective spectral normalization by stretching the data along the axis of small variances and squeezing the data in the direction with large variances, while the inverse square root transforms the data into the uncorrelated structure that has unit variance in all directions.}
    \label{fig:cover}
\end{figure}

Consider a positive semi-definite matrix $\mA$. The principle square root $\mA^{\frac{1}{2}}$ and the inverse square root $\mA^{-\frac{1}{2}}$ are mathematically of practical interests, mainly because some desired spectral properties can be obtained by such transformations. An exemplary illustration is given in Fig.~\ref{fig:cover}. As can be seen, the matrix square root can shrink/stretch the feature variances along with the direction of principle components, which is known as an effective spectral normalization for covariance matrices. The inverse square root, on the other hand, can be used to whiten the data, \emph{i.e.,} make the data has a unit variance in each dimension. These appealing spectral properties are very useful in many computer vision applications. In Global Covariance Pooling (GCP)~\cite{lin2017improved,li2017second,li2018towards,song2021approximate} and other related high-order representation methods~\cite{xie2021so,gao2021temporal}, the matrix square root is often used to normalize the high-order feature, which can benefit some classification tasks like general visual recognition~\cite{li2017second,li2018towards,xie2021so}, fine-grained visual categorization~\cite{song2022eigenvalues}, and video action recognition~\cite{gao2021temporal}. The inverse square root is used as the whitening transform to eliminate the feature correlation, which is widely applied in decorrelated Batch Normalization (BN)~\cite{huang2018decorrelated,huang2019iterative,huang2020investigation} and other related models that involve the whitening transform~\cite{siarohin2018whitening,ermolov2021whitening}. In the field of neural style transfer, both the matrix square root and its inverse are adopted to perform successive Whitening and Coloring Transform (WCT) to transfer the style information for better generation fidelity~\cite{li2017universal,cho2019image,choi2021robustnet}. 



To compute the matrix square root, the standard method is via Singular Value Decomposition (SVD). Given the real symmetric matrix $\mA$, its matrix square root is computed as:
\begin{equation}
    \mA^{\frac{1}{2}} =(\mU\mLambda\mU^{T})^{\frac{1}{2}} = \mU\mLambda^{\frac{1}{2}}\mU^{T}
\end{equation}
where $\mU$ is the eigenvector matrix, and $\mLambda$ is the diagonal eigenvalue matrix. As derived by Ionescu~\emph{et al.}~\cite{ionescu2015training}, the partial derivative of the eigendecomposition is calculated as:
\begin{equation}
    \frac{\partial l}{\partial \mA} = \mU\Big(\mK^{T}\odot(\mU^{T}\frac{\partial l}{\partial\mU})+(\frac{\partial l}{\partial\mLambda})_{\rm diag}\Big)\mU^{T}
    \label{svd_back}
\end{equation}
where $l$ is the loss function, $\odot$ denotes the element-wise product, and $()_{\rm diag}$ represents the operation of setting the off-diagonal entries to zero. Despite the long-studied theories and well-developed algorithms of SVD, there exist two obstacles when integrating it into deep learning frameworks. One issue is the back-propagation instability. For the matrix $\mK$ defined in~\cref{svd_back}, its off-diagonal entry is $K_{ij}{=}\nicefrac{1}{(\lambda_{i}-\lambda_{j})}$, where $\lambda_{i}$ and $\lambda_{j}$ are involved eigenvalues. When the two eigenvalues are close and small, the gradient is very likely to explode, \emph{i.e.,} $K_{ij}{\rightarrow}{\infty}$. This issue has been solved by some methods that use approximation techniques to estimate the gradients~\cite{wang2019backpropagation,wang2021robust,song2021approximate}. The other problem is the expensive time cost of the forward eigendecomposition. As the SVD is not supported well by GPUs~\cite{lahabar2009singular}, performing the eigendecomposition on the deep learning platforms is rather time-consuming. Incorporating the SVD with deep models could add extra burdens to the training process. Particularly for batched matrices, modern deep learning frameworks, such as Tensorflow and Pytorch, give limited optimization for the matrix decomposition within the mini-batch. They inevitably use a for-loop to conduct the SVD one matrix by another. However, how to efficiently perform the SVD in the context of deep learning has not been touched by the research community.  


To avoid explicit eigendecomposition, one commonly used alternative is the Newton-Schulz iteration (NS iteration)~\cite{schulz1933iterative,higham2008functions} which  modifies the ordinary Newton iteration by replacing the matrix inverse but preserving the quadratic convergence. Compared with SVD, the NS iteration is rich in matrix multiplication and more GPU-friendly. Thus, this technique has been widely used to approximate the matrix square root in different applications~\cite{lin2017improved,li2018towards,huang2019iterative}. The forward computation relies on the following coupled iterations:
\begin{equation}
    \mY_{k+1}=\frac{1}{2}\mY_{k} (3\mI - \mZ_{k}\mY_{k}), \mZ_{k+1}=\frac{1}{2}(3\mI-\mZ_{k}\mY_{k})\mZ_{k}
    \label{eq:ns_fp}
\end{equation}
where $\mY_{k}$ and $\mZ_{k}$ converge to $\mA^{\frac{1}{2}}$ and $\mA^{-\frac{1}{2}}$, respectively. Since the NS iteration only converges locally (\emph{i.e.,} $||\mA||_{2}{<}1$), we need to pre-normalize the initial matrix and post-compensate the resultant approximation as $\mY_{0}{=}\frac{1}{||\mA||_{\rm F}}\mA$ and$\ \mA^{\frac{1}{2}}{=}\sqrt{||\mA||_{\rm F}}\mY_{k}$. Each forward iteration involves $3$ matrix multiplications, which is more efficient than the forward pass of SVD. However, the backward pass of the NS iteration takes $14$ matrix multiplications per iteration. Consider that the NS iteration often takes $5$ iterations to achieve reasonable performances~\cite{li2018towards,huang2019iterative}. The backward pass is much more time-costing than the backward algorithm of SVD. The speed improvement could be larger if a more efficient backward algorithm is developed.

To address the drawbacks of SVD and NS iteration, \emph{i.e.} the low efficiency in either the forward or backward pass, we derive two methods \textbf{that are efficient in both forward and backward propagation} to compute the differentiable matrix square root and its inverse. In the forward pass (FP), we propose using Matrix Taylor Polynomial (MTP) and Matrix Pad\'e Approximants (MPA) for approximating the matrix square root. The former approach is slightly faster but the latter is more numerically accurate. Both methods yield considerable speed-up compared with the SVD or the NS iteration in the forward computation. The proposed MTP and MPA can be also used to approximate the inverse square root without any additional computational cost. For the backward pass (BP), we consider the gradient function as a Lyapunov equation and propose an iterative solution using the matrix sign function. The backward pass costs fewer matrix multiplications and is more computationally efficient than the NS iteration. Our proposed iterative Lyapunov solver applies to both the matrix square root and the inverse square root. The only difference is that deriving the gradient of inverse square root requires $3$ more matrix multiplications than computing that of matrix square root. 

Through a series of numerical tests, we show that the proposed MTP-Lya and MPA-Lya deliver consistent speed improvement for different batch sizes, matrix dimensions, and some hyper-parameters (\emph{e.g.,} degrees of power series to match and iteration times). Moreover, our proposed MPA-Lya consistently gives a better approximation of the matrix square root and its inverse than the NS iteration. Besides the numerical tests, we conduct extensive experiments in a number of computer vision applications, including decorrelated batch normalization, second-order vision transformer, global covariance pooling for large-scale and fine-grained image recognition, attentive global covariance pooling for video action recognition, and neural style transfer. Our methods can achieve competitive performances against the SVD and the NS iteration with the least amount of time overhead. Our MPA is suitable in use cases where the high precision is needed, while our MTP works in applications where the accuracy is less demanded but the efficiency is more important. The contributions of the paper are twofold:

\begin{itemize}
    \item We propose two fast methods that compute the differentiable matrix square root and the inverse square root. The forward propagation relies on the matrix Taylor polynomial or matrix Pad\'e approximant, while an iterative backward gradient solver is derived from the Lyapunov equation using the matrix sign function. 
    \item Our proposed algorithms are validated by a series of numerical tests and several real-world computer vision applications. The experimental results demonstrate that our methods have a faster calculation speed and also have very competitive performances. 
\end{itemize}

This paper is an expanded version of~\cite{song2022fast}. In the conference paper~\cite{song2022fast}, the proposed fast algorithms only apply to the matrix square root $\mA^{\frac{1}{2}}$. For the application of inverse square root $\mA^{-\frac{1}{2}}$, we have to solve the linear system or compute the matrix inverse. However, both techniques are not GPU-efficient enough and could add extra computational burdens to the training. In this extended manuscript, we target the drawback and extend our algorithm to the case of inverse square root, which avoids the expensive computation and allows for faster calculation in more application scenarios. Compared with computing the matrix square root, computing the inverse square root consumes the same time complexity in the FP and requires 3 more matrix multiplications in the BP. The paper thus presents a complete solution to the efficiency issue of the differentiable spectral layer. Besides the algorithm extension, our method is validated in more computer vision applications: global covariance pooling for image/video recognition and neural style transfer. We also shed light on the peculiar incompatibility of NS iteration and Lyapunov solver discussed in Sec.~\ref{sec:lya_backward}.


The rest of the paper is organized as follows: Sec.~\ref{sec:related} describes the computational methods and applications of differentiable matrix square root and its inverse. Sec.~\ref{sec:method} introduces our method that computes the end-to-end matrix square root, and Sec.~\ref{sec:method_inverse} presents the extension of our method to the inverse square root. Sec.~\ref{sec:exp} provides the experimental results, the ablation studies, and some in-depth analysis. Finally, Sec.~\ref{sec:conclusion} summarizes the conclusions.

\section{Related Work}\label{sec:related}

In this section, we recap the previous approaches that compute the differentiable matrix square root and the inverse square root, followed by a discussion on the usage in some applications of deep learning and computer vision.

\subsection{Computational Methods}
Ionescu~\emph{et al.}~\cite{ionescu2015training,ionescu2015matrix} first formulate the theory of matrix back-propagation, making it possible to integrate a spectral meta-layer into neural networks. Existing approaches that compute the differentiable matrix square root and its inverse are mainly based on the SVD or NS iteration. The SVD calculates the accurate solution but suffers from backward instability and expensive time cost, whereas the NS iteration computes the approximate solution but is more GPU-friendly. For the backward algorithm of SVD, several methods have been proposed to resolve this gradient explosion issue~\cite{wang2019backpropagation,Dang18a,Dang20a,wang2021robust,song2021approximate}. Wang~\emph{et al.}~\cite{wang2019backpropagation} propose to apply Power Iteration (PI) to approximate the SVD gradient. Recently, Song~\emph{et al.}~\cite{song2021approximate} propose to rely on Pad\'e approximants to closely estimate the backward gradient of SVD. 

To avoid explicit eigendecomposition, Lin~\emph{et al.}~\cite{lin2017improved} propose to substitute SVD with the NS iteration. Following this work, Li~\emph{et al.}~\cite{li2017second} and Huang~\emph{et al.}~\cite{huang2018decorrelated} adopt the NS iteration in the task of global covariance pooling and decorrelated batch normalization, respectively. For the backward pass of the differentiable matrix square root, Lin~\emph{et al.}~\cite{lin2017improved} also suggest viewing the gradient function as a Lyapunov equation. However, their proposed exact solution is infeasible to compute practically, and the suggested Bartels-Steward algorithm~\cite{bartels1972solution} requires explicit eigendecomposition or Schur decomposition, which is again not GPU-friendly. By contrast, our proposed iterative solution using the matrix sign function is more computationally efficient and achieves comparable performances against the Bartels-Steward algorithm (see the ablation study in Sec.~\ref{sec:abla}).

\subsection{Applications}

\subsubsection{Global Covariance Pooling}

One successful application of the differentiable matrix square root is the Global Covariance Pooling (GCP), which is a meta-layer inserted before the FC layer of deep models to compute the matrix square root of the feature covariance. Equipped with the GCP meta-layers, existing deep models have achieved state-of-the-art performances on both generic and fine-grained visual recognition ~\cite{lin2015bilinear,li2017second,lin2017improved,li2018towards,wang2019deep,wang2020deep,song2021approximate,song2022eigenvalues}. Inspired by recent advances of transformers~\cite{vaswani2017attention}, Xie~\emph{et al.}~\cite{xie2021so} integrate the GCP meta-layer into the vision transformer~\cite{dosovitskiy2020image} to exploit the second-order statistics of the high-level visual tokens, which solves the issue that vision transformers need pre-training on ultra-large-scale datasets. More recently, Gao~\emph{et al.}~\cite{gao2021temporal} propose an attentive and temporal-based GCP model for video action recognition. 

\subsubsection{Decorrelated Batch Normalization}

Another line of research proposes to use ZCA whitening, which applies the inverse square root of the covariance to whiten the feature, as an alternative scheme for the standard batch normalization~\cite{ioffe2015batch}. The whitening procedure, \emph{a.k.a} decorrelated batch normalization, does not only standardize the feature but also eliminates the data correlation. The decorrelated batch normalization can improve both the optimization efficiency and generalization ability of deep neural networks~\cite{huang2018decorrelated,siarohin2018whitening,huang2019iterative,pan2019switchable,huang2020investigation,ermolov2021whitening,huang2021group,zhang2021stochastic,cho2021improving}. 

\subsubsection{Whitening and Coloring Transform}

The WCT~\cite{li2017universal} is also an active research field where the differentiable matrix square root and its inverse are widely used. In general, the WCT performs successively the whitening transform (using inverse square root) and the coloring transform (using matrix square root) on the multi-scale features to preserve the content of current image but carrying the style of another image. During the past few years, the WCT methods have achieved remarkable progress in universal style transfer~\cite{li2017universal,li2018closed,wang2020diversified}, domain adaptation~\cite{abramov2020keep,choi2021robustnet}, and image translation~\cite{ulyanov2017improved,cho2019image}.

Besides the three main applications discussed above, there are still some minor applications, such as semantic segmentation~\cite{sun2021second} and super resolution~\cite{Dai_2019_CVPR}. 
\begin{table}[htbp]
\caption{Summary of mathematical notation and symbol.}
    \centering
    \begin{tabular}{c|c}
    \toprule
        $\mA^{p}$ & Matrix $p$-th power.\\
        $\mI$ & Identity matrix. \\
        $||\cdot||_{\rm F}$ & Matrix Frobenius norm. \\
        $\dbinom{n}{k}$ & Binomial coefficients calculated as $\nicefrac{n!}{k!(n-k)!}$.\\
        $vec(\cdot)$ & Unrolling matrix into vector.\\ 
        $\otimes$ & Matrix Kronecker product. \\
        $sign(\mA)$ & Matrix sign function calculated as $\mA(\mA^{2})^{-\frac{1}{2}}$\\
        $\frac{\partial l}{\partial \mA}$ & Partial derivative of loss $l$ w.r.t. matrix $\mA$\\
    \bottomrule
    \end{tabular}
    \label{tab:notation}
\end{table}

\section{Fast Differentiable Matrix Square Root}\label{sec:method}

Table~\ref{tab:notation} summarizes the notation we will use from now on. This section presents the forward pass and the backward propagation of our fast differentiable matrix square root. For the inverse square root, we introduce the derivation in Sec.~\ref{sec:method_inverse}.


\subsection{Forward Pass}

\subsubsection{Matrix Taylor Polynomial}

We begin with motivating the Taylor series for the scalar case. Consider the following power series:
\begin{equation}
    (1-z)^{\frac{1}{2}} = 1 - \sum_{k=1}^{\infty} \Big|\dbinom{\frac{1}{2}}{k}\Big| z^{k}
    \label{taylor_scalar}
\end{equation}
where $\dbinom{\frac{1}{2}}{k}$ denotes the binomial coefficients that involve fractions, and the series converges when $z{<}1$ according to the Cauchy root test. For the matrix case, the power series can be similarly defined by:
\begin{equation}
    (\mI-\mZ)^{\frac{1}{2}}=\mI - \sum_{k=1}^{\infty} \Big|\dbinom{\frac{1}{2}}{k}\Big|  \mZ^{k}
    \label{taylor_matrix}
\end{equation}
where $\mI$ is the identity matrix. Let us substitute $\mZ$ with $(\mI{-}\mA)$, we can obtain:
\begin{equation}
    \mA^{\frac{1}{2}} = \mI - \sum_{k=1}^{\infty} \Big|\dbinom{\frac{1}{2}}{k}\Big|  (\mI-\mA)^k
    \label{mtp_unnorm}
\end{equation}
Similar with the scalar case, the power series converge only if $||(\mI-\mA)||_{p}{<}1$, where $||\cdot||_{p}$ denotes any vector-induced matrix norms. To circumvent this issue, we can first pre-normalize the matrix $\mA$ by dividing $||\mA||_{\rm F}$. This can guarantee the convergence as $||\mI{-}\frac{\mA}{||\mA||_{\rm F}}||_{p}{<}1$ is always satisfied. Afterwards, the matrix square root $\mA^{\frac{1}{2}}$ is post-compensated by multiplying $\sqrt{||\mA||_{\rm F}}$. Integrated with these two operations, \cref{mtp_unnorm} can be re-formulated as:
\begin{equation}
    \mA^{\frac{1}{2}} =\sqrt{||\mA||_{\rm F}}\cdot \Big(\mI - \sum_{k=1}^{\infty} \Big|\dbinom{\frac{1}{2}}{k}\Big| (\mI-\frac{\mA}{||\mA||_{\rm F}})^{k} \Big)
    \label{mtp_square}
\end{equation}
Truncating the series to a certain degree $K$ yields the MTP approximation for the matrix square root. For the MTP of degree $K$, $K{-}1$ matrix multiplications are needed.


\subsubsection{Matrix Pad\'e Approximant}

\begin{figure}[htbp]
    \centering
    \includegraphics[width=0.9\linewidth]{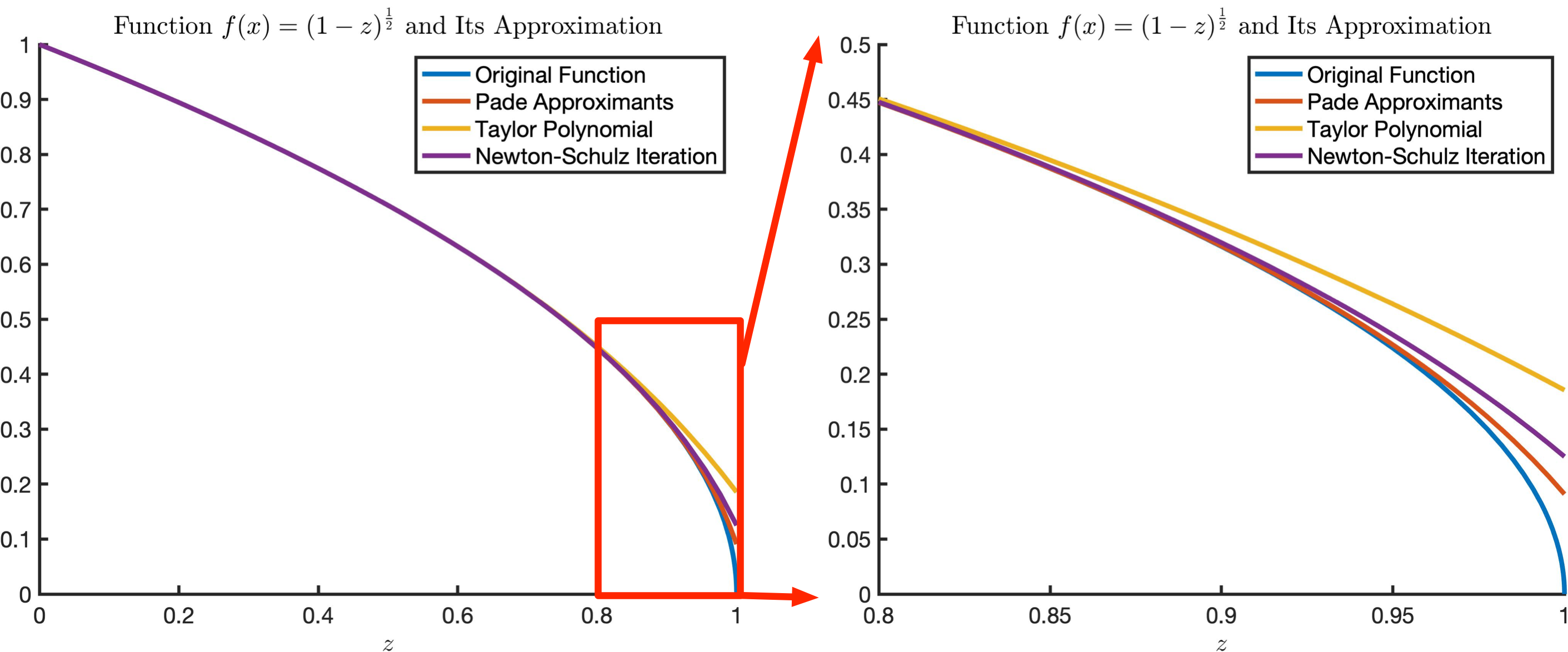}
    \caption{The function $(1-z)^{\frac{1}{2}}$ in the range of $|z|<1$ and its approximation including Taylor polynomial, Newton-Schulz iteration, and Pad\'e approximants. The Pad\'e approximants consistently achieves a better estimation for other approximation schemes for any possible input values. }
    \label{fig:approx_mpa_mtp}
\end{figure}

The MTP enjoys the fast calculation, but it converges uniformly and sometimes suffers from the so-called "hump phenomenon", \emph{i.e.,} the intermediate terms of the series grow quickly but cancel each other in the summation, which results in a large approximation error. Expanding the series to a higher degree does not solve this issue either. The MPA, which adopts two polynomials of smaller degrees to construct a rational approximation, is able to avoid this caveat. To visually illustrate this impact, we depict the approximation of the scalar square root in Fig.~\ref{fig:approx_mpa_mtp}. The Pad\'e approximants consistently deliver a better approximation than NS iteration and Taylor polynomial. In particular, when the input is close to the convergence boundary ($z{=}1$) where NS iteration and Taylor polynomials suffer from a larger approximation error, our Pad\'e approximants still present a reasonable estimation. The superior property also generalizes to the matrix case. 

\begin{figure}
    \centering
    \includegraphics[width=0.7\linewidth]{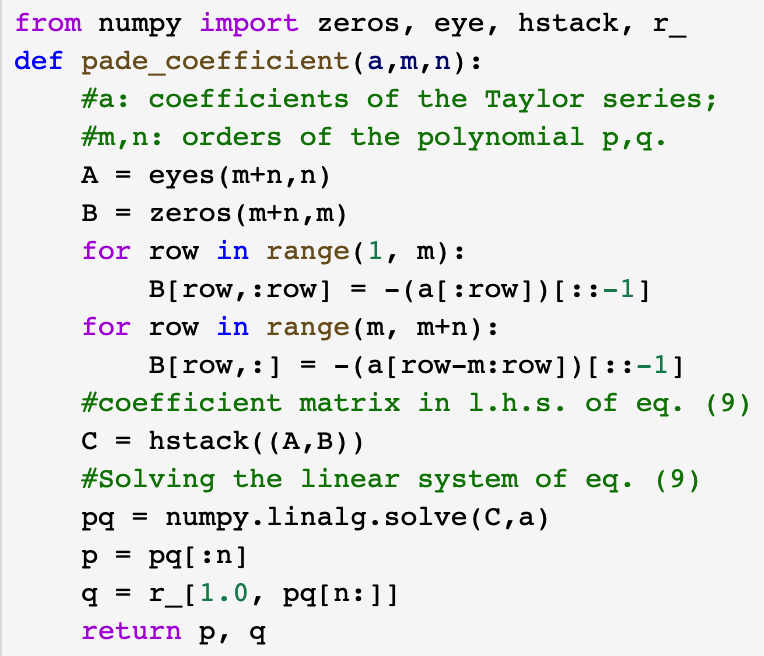}
    \caption{Python-like pseudo-codes for Pad\'e coefficients.}
    \label{fig:pade_code}
\end{figure}


The MPA is computed as the fraction of two sets of polynomials: denominator polynomial $\sum_{n=1}^{N}q_{n}z^{n}$ and numerator polynomial $\sum_{m=1}^{M}p_{m}z^{m}$. The coefficients $q_{n}$ and $p_{m}$ are pre-computed by matching to the corresponding Taylor series. Given the power series of scalar in~\cref{taylor_scalar}, the coefficients of a $[M,N]$ scalar Pad\'e approximant are computed by matching to the series of degree $M{+}N{+}1$:
\begin{equation}
    \frac{1-\sum_{m=1}^{M}p_{m}z^{m}}{1-\sum_{n=1}^{N}q_{n}z^{n}}  = 1 - \sum_{k=1}^{M+N} \Big|\dbinom{\frac{1}{2}}{k}\Big| z^{k}
    \label{pade_match}
\end{equation}
where $p_{m}$ and $q_{n}$ also apply to the matrix case. This matching gives rise to a system of linear equations:
\begin{equation}
\begin{cases}
     - \Big|\dbinom{\frac{1}{2}}{1}\Big| - q_{1} = -p_{1},\\
     - \Big|\dbinom{\frac{1}{2}}{2}\Big| + \Big|\dbinom{\frac{1}{2}}{1}\Big|q_{1} - q_{2} = -p_{2}, \\
     - \Big|\dbinom{\frac{1}{2}}{M}\Big| + \Big|\dbinom{\frac{1}{2}}{M-1}\Big|q_{1} + \cdots -q_{M} = p_{M}, \\
     \cdots\cdots
\end{cases} 
\label{eq:pade_linearsystem}
\end{equation}
Solving these equations directly determines the coefficients. We give the Python-like pseudo-codes in Fig.~\ref{fig:pade_code}. The numerator polynomial and denominator polynomials of MPA are given by:
\begin{equation}
    \begin{gathered}
    \mP_{M}= \mI - \sum_{m=1}^{M} p_{m} (\mI-\frac{\mA}{||\mA||_{\rm F}})^{m},\\ 
    \mQ_{N}= \mI - \sum_{n=1}^{N} q_{n} (\mI-\frac{\mA}{||\mA||_{\rm F}})^{n}.
    \end{gathered}
    \label{ration_app}
\end{equation}
Then the MPA for approximating the matrix square root is computed as:
\begin{equation}
    \mA^{\frac{1}{2}} = \sqrt{||\mA||_{\rm F}}\mQ_{N}^{-1}\mP_{M}.
    \label{mpa_square}
\end{equation}
Compared with the MTP, the MPA trades off half of the matrix multiplications with one matrix inverse, which slightly increases the computational cost but converges more quickly and delivers better approximation abilities. Moreover, we note that the matrix inverse can be avoided, as \cref{mpa_square} can be more efficiently and numerically stably computed by solving the linear system $\mQ_{N}\mA^{\frac{1}{2}}{=} \sqrt{||\mA||_{\rm F}}\mP_{M}$.
According to Van~\emph{et al.}~\cite{van2006pade}, diagonal Pad\'e approximants (\emph{i.e.,} $\mP_{M}$ and $\mQ_{N}$ have the same degree) usually yield better approximation than the non-diagonal ones. Therefore, to match the MPA and MTP of the same degree, we set $M{=}N{=}\frac{K-1}{2}$.

\begin{table}
\centering
\caption{Comparison of forward operations. For the matrix square root and its inverse, our MPA/MTP consumes the same complexity. The cost of $1$ NS iteration is about that of MTP of $4$ degrees and about that of MPA of $2$ degrees.}\label{tab:forward_complexity}
\begin{tabular}{c|c|c|c}
    \hline
     Op.    & MTP & MPA & NS iteration \\
     \hline
    Mat. Mul.  & $K{-}1$ & $\nicefrac{(K{-}1)}{2}$ & 3 $\times$ \#iters \\
    Mat. Inv.  & 0 & 1        & 0 \\
    \hline
\end{tabular}
\end{table}

Table~\ref{tab:forward_complexity} summarizes the forward computational complexity. As suggested in Li~\emph{et al.}~\cite{li2018towards} and Huang~\emph{et al.}~\cite{huang2019iterative}, the iteration times for NS iteration are often set as $5$ such that reasonable performances can be achieved. That is, to consume the same complexity as the NS iteration does, our MTP and MPA can match to the power series up to degree $16$. However, as illustrated in Fig.~\ref{fig:fp_sped_err}, our MPA achieves better accuracy than the NS iteration even at degree $8$. This observation implies that our MPA is a better option in terms of both accuracy and speed. 

\subsection{Backward Pass}

Though one can manually derive the gradient of the MPA and MTP, their backward algorithms are computationally expensive as they involve the matrix power up to degree $K$, where $K$ can be arbitrarily large. Relying on the AutoGrad package of deep learning frameworks can be both time- and memory-consuming since the gradients of intermediate variables would be computed and the matrix inverse of MPA is involved. To attain a more efficient backward algorithm, we propose to iteratively solve the gradient equation using the matrix sign function. Given the matrix $\mathbf{A}$ and its square root $\mA^{\frac{1}{2}}$, since we have $\mA^{\frac{1}{2}}\mA^{\frac{1}{2}}{=}\mA$, a perturbation on $\mA$ leads to:
\begin{equation}
    \mA^{\frac{1}{2}} d \mA^{\frac{1}{2}} + d\mA^{\frac{1}{2}} \mA^{\frac{1}{2}} = d \mA
\end{equation}
Using the chain rule, the gradient function of the matrix square root satisfies:
\begin{equation}
    \mA^{\frac{1}{2}} \frac{\partial l}{\partial \mA} + \frac{\partial l}{\partial \mA} \mA^{\frac{1}{2}} = \frac{\partial l}{\partial \mA^{\frac{1}{2}}}
    \label{gradient_function}
\end{equation}
As pointed out by Li~\emph{et al.}~\cite{lin2017improved}, \cref{gradient_function} actually defines the continuous-time Lyapunov equation ($\mB\mX {+} \mX\mB {=} \mC$) or a special case of Sylvester equation ($\mB\mX {+} \mX\mD {=} \mC$). The closed-form solution is given by:
\begin{equation}
    vec(\frac{\partial l}{\partial \mA}) = \Big(\mA^{\frac{1}{2}}\otimes \mI + \mI \otimes \mA^{\frac{1}{2}} \Big)^{-1} vec(\frac{\partial l}{\partial \mA^{\frac{1}{2}}})
\end{equation}
where $vec(\cdot)$ denotes unrolling a matrix to vectors, and $\otimes$ is the Kronecker product. Although the closed-form solution exists theoretically, it cannot be computed in practice due to the huge memory consumption of the Kronecker product. Supposing that both $\mA^{\frac{1}{2}}$ and $\mI$ are of size $256{\times}256$, the Kronecker product $\mA^{\frac{1}{2}}{\otimes}\mI$ would take the dimension of $256^{2}{\times}256^{2}$, which is infeasible to compute or store.
Another approach to solve~\cref{gradient_function} is via the Bartels-Stewart algorithm~\cite{bartels1972solution}. However, it requires explicit eigendecomposition or Schulz decomposition, which is not GPU-friendly and computationally expensive. 

To attain a GPU-friendly gradient solver, we propose to use the matrix sign function and iteratively solve the Lyapunov equation. Solving the Sylvester equation via matrix sign function has been long studied in the literature of numerical analysis~\cite{roberts1980linear,kenney1995matrix,benner2006solving}. One notable line of research is using the family of Newton iterations. Consider the following continuous Lyapunov function:
\begin{equation}
    \mB\mX + \mX\mB = \mC
    \label{lyapunov}
\end{equation}
where $\mB$ refers to $\mA^{\frac{1}{2}}$ in~\cref{gradient_function}, $\mC$ represents $\frac{\partial l}{\partial \mA^{\frac{1}{2}}}$, and $\mX$ denotes the seeking solution $\frac{\partial l}{\partial \mA}$. Eq.~(\ref{lyapunov}) can be represented by the following block using a Jordan decomposition:
\begin{equation}
    \mH=\begin{bmatrix}
    \mB & \mC\\
    \mathbf{0} & -\mB
    \end{bmatrix} = 
    \begin{bmatrix}
    \mI & \mX\\
    \mathbf{0} & \mI
    \end{bmatrix} 
    \begin{bmatrix}
    \mB & \mathbf{0}\\
    \mathbf{0} & -\mB
    \end{bmatrix} 
    \begin{bmatrix}
    \mI & \mX\\
    \mathbf{0} & \mI
    \end{bmatrix}^{-1} 
    \label{block_lyapunov}
\end{equation}
The matrix sign function is invariant to the Jordan canonical form or spectral decomposition. This property allows the use of Newton's iterations for iteratively solving the Lyapunov function. Specifically, we have:
\begin{lemma}[Matrix Sign Function~\cite{higham2008functions}]
  \label{sign_1}
  For a given matrix $\mH$ with no eigenvalues on the imaginary axis, its sign function has the following properties: 1) $sign(\mH)^2=\mI$; 2) if $\mH$ has the Jordan decomposition $\mH{=}\mT\mM\mT^{-1}$, then its sign function satisfies $sign(\mH){=}\mT  sign(\mM) \mT^{-1}$.
\end{lemma}
We give the complete proof in the Supplementary Material. Lemma~\ref{sign_1}.1 shows that $sign(\mH)$ is the matrix square root of the identity matrix, which indicates the possibility of using Newton's root-finding method to derive the solution~\cite{higham2008functions}. Here we also adopt the Newton-Schulz iteration, the modified inverse-free and multiplication-rich Newton iteration, to iteratively compute $sign(\mH)$. This leads to the coupled iteration as:
\begin{equation}
\begin{gathered}
     \mB_{k+1} =\frac{1}{2} \mB_{k} (3\mI-\mB_{k}^2), \\
     \mC_{k+1} =\frac{1}{2} \Big(-\mB_{k}^{2}\mC_{k} + \mB_{k}\mC_{k}\mB_{k} + \mC_{k}(3\mI-\mB_{k}^2)\Big).
     \label{lya_iterations}
\end{gathered}
\end{equation}
The equation above defines two coupled iterations for solving the Lyapunov equation. Since the NS iteration converges only locally, \emph{i.e.,} converges when $||\mH_{k}^{2}{-}\mI||{<}1$, here we divide $\mH_{0}$ by $||\mB||_{\rm F}$ to meet the convergence condition. This normalization defines the initialization $\mB_{0}{=}\frac{\mB}{||\mB||_{\rm F}}$ and $\mC_{0}{=}\frac{\mC}{||\mB||_{\rm F}}$. Relying on Lemma~\ref{sign_1}.2, the sign function of \cref{block_lyapunov} can be also calculated as:
\begin{equation}
\begin{aligned}
    sign(\mH)&=
     sign\Big(\begin{bmatrix}
    \mB & \mC\\
    \mathbf{0} & -\mB
    \end{bmatrix}\Big)=\begin{bmatrix}
    \mI & 2 \mX\\
    \mathbf{0} & -\mI
    \end{bmatrix} 
\end{aligned}
\label{sign_block_lya}
\end{equation}

As indicated above, the iterations in~\cref{lya_iterations} have the convergence:
\begin{equation}
    \lim_{k\rightarrow\infty}\mB_{k} = \mathbf{I}, \lim_{k\rightarrow\infty}\mC_{k} = 2\mX
\end{equation}
After iterating $k$ times, we can get the approximate solution $\mX{=}\frac{1}{2}\mC_{k}$. Instead of choosing setting iteration times, one can also set the termination criterion by checking the convergence $||\mB_{k}-\mI||_{\rm F}{<}\tau$, where $\tau$ is the pre-defined tolerance. 

Table~\ref{tab:backward_complexiy} compares the backward computation complexity of the iterative Lyapunov solver and the NS iteration. Our proposed Lyapunov solver spends fewer matrix multiplications and is thus more efficient than the NS iteration. Even if we iterate the Lyapunov solver more times (\emph{e.g.,} 7 or 8), it still costs less time than the backward calculation of NS iteration that iterates $5$ times.

\begin{table}
\centering
\caption{Comparison of backward operations. For the inverse square root, our Lyapunov solver uses marginally $3$ more matrix multiplications. The cost of $1$ NS iteration is about that of $2$ iterations of Lyapunov solver.}\label{tab:backward_complexiy}
\begin{tabular}{c|c|c|c}
    \hline
     Op.    &  Lya (Mat. Sqrt.)  &  Lya (Inv. Sqrt.) & NS iteration\\
     \hline
    Mat. Mul.  & 6 $\times$ \#iters & 3 + 6 $\times$ \#iters& 4 + 10 $\times$ \#iters \\
    Mat. Inv. & 0   & 0  & 0 \\
    \hline
    \end{tabular}
\end{table}

\section{Fast Differentiable Inverse Square Root}\label{sec:method_inverse}

In this section, we introduce the extension of our algorithm to the inverse square root.

\subsection{Forward Pass}

\subsubsection{Matrix Taylor Polynomial}

To derive the MTP of inverse square root, we need to match to the following power series:
\begin{equation}
    (1-z)^{-\frac{1}{2}} = 1 + \sum_{k=1}^{\infty}\Big|\dbinom{-\frac{1}{2}}{k}\Big| z^{k}
    \label{taylor_inv_scalar}
\end{equation}
Similar with the procedure of the matrix square root in~\cref{taylor_matrix,mtp_unnorm}, the MTP approximation can be computed as:
\begin{equation}
    \mA^{-\frac{1}{2}} =\mI + \sum_{k=1}^{\infty} \Big|\dbinom{-\frac{1}{2}}{k}\Big| (\mI-\frac{\mA}{||\mA||_{\rm F}})^{k}
\end{equation}
Instead of the post-normalization of matrix square root by multiplying $\sqrt{||\mA||_{\rm F}}$ as done in~\cref{mtp_square}, we need to divide $\sqrt{||\mA||_{\rm F}}$ for computing the inverse square root:
\begin{equation}
    \mA^{-\frac{1}{2}} =\frac{1}{\sqrt{||\mA||_{\rm F}}}\cdot \Big(\mI + \sum_{k=1}^{\infty} \Big|\dbinom{-\frac{1}{2}}{k}\Big| (\mI-\frac{\mA}{||\mA||_{\rm F}})^{k} \Big)
\end{equation}
Compared with the MTP of matrix square root in the same degree, the inverse square root consumes the same computational complexity. 



\subsubsection{Matrix Pad\'e Approximant}

The matrix square root $\mA^{\frac{1}{2}}$ of our MPA is calculated as $\sqrt{||\mA||_{\rm F}}\mQ_{N}^{-1}\mP_{M}$. For the inverse square root, we can directly compute the inverse as:
\begin{equation}
    \mA^{-\frac{1}{2}}=(\sqrt{||\mA||_{\rm F}}\mQ_{N}^{-1}\mP_{M})^{-1}=\frac{1}{\sqrt{||\mA||_{\rm F}}}\mP_{M}^{-1}\mQ_{N}
    \label{mpa_square_inverse1}
\end{equation}
The extension to inverse square root comes for free as it does not require additional computation. For both the matrix square root and inverse square root, the matrix polynomials $\mQ_{N}$ and $\mP_{M}$ need to be first computed, and then one matrix inverse or solving the linear system is required. 

Another approach to derive the MPA for inverse square root is to match the power series in~\cref{taylor_inv_scalar} and construct the MPA again. The matching is calculated as:
\begin{equation}
    \frac{1+\sum_{m=1}^{M}r_{m}z^{m}}{1+\sum_{n=1}^{N}s_{n}z^{n}}  = 1 + \sum_{k=1}^{M+N} \Big|\dbinom{-\frac{1}{2}}{k}\Big| z^{k}
\end{equation}
where $r_{m}$ and $s_{n}$ denote the new Pad\'e coefficients. Then the matrix polynomials are computed as:
\begin{equation}
    \begin{gathered}
    \mR_{M}= \mI + \sum_{m=1}^{M} r_{m} (\mI-\frac{\mA}{||\mA||_{\rm F}})^{m},\\ 
    \mS_{N}= \mI + \sum_{n=1}^{N} s_{n} (\mI-\frac{\mA}{||\mA||_{\rm F}})^{n}.
    \end{gathered}
\end{equation}
The MPA for approximating the inverse square root is calculated as:
\begin{equation}
    \mA^{-\frac{1}{2}} = \frac{1}{\sqrt{||\mA||_{\rm F}}}\mS_{N}^{-1}\mR_{M}.
    \label{mpa_square_inverse2}
\end{equation}
This method for deriving MPA also leads to the same complexity. Notice that these two different computation methods are equivalent to each other. Specifically, we have:

\begin{prop}
The diagonal MPA $\frac{1}{\sqrt{||\mA||_{\rm F}}}\mS_{N}^{-1}\mR_{M}$ is equivalent to the diagonal MPA $\frac{1}{\sqrt{||\mA||_{\rm F}}}\mP_{M}^{-1}\mQ_{N}$, and the relation $p_{m}{=}-s_{n}$ and $q_{n}{=}-r_{m}$ hold for any $m{=}n$.
\end{prop}

We give the detailed proof in Supplementary Material. Since two sets of MPA are equivalent, we adopt the implementation of inverse square root in~\cref{mpa_square_inverse1} throughout our experiments, as it shares the same $\mP_{M}$ and $\mQ_{N}$ with the matrix square root.

\subsection{Backward Pass}

For the inverse square root, we can also rely on the iterative Lyapunov solver for the gradient computation. Consider the following relation:
\begin{equation}
    \mA^{\frac{1}{2}}\mA^{-\frac{1}{2}}=\mI. 
\end{equation}
A perturbation on both sides leads to:
\begin{equation}
     d\mA^{\frac{1}{2}}\mA^{-\frac{1}{2}} + \mA^{\frac{1}{2}}d\mA^{-\frac{1}{2}} = d\mI.
\end{equation}
Using the chain rule, we can obtain the gradient equation after some arrangements:
\begin{equation}
    \frac{\partial l}{\partial \mA^{\frac{1}{2}}} = -\mA^{-\frac{1}{2}}\frac{\partial l}{\partial \mA^{-\frac{1}{2}}}\mA^{-\frac{1}{2}}.
\end{equation}
Injecting this equation into~\cref{gradient_function} leads to the re-formulation:
\begin{equation}
\begin{gathered}
     \mA^{\frac{1}{2}} \frac{\partial l}{\partial \mA} + \frac{\partial l}{\partial \mA} \mA^{\frac{1}{2}} = -\mA^{-\frac{1}{2}}\frac{\partial l}{\partial \mA^{-\frac{1}{2}}}\mA^{-\frac{1}{2}} \\
   \mA^{-\frac{1}{2}}\frac{\partial l}{\partial \mA} + \frac{\partial l}{\partial \mA}\mA^{-\frac{1}{2}} = -\mA^{-1}\frac{\partial l}{\partial \mA^{-\frac{1}{2}}}\mA^{-1}.
    \label{inverse_gradient}
\end{gathered}
\end{equation}
As can be seen, now the gradient function resembles the continuous Lyapunov equation again. The only difference with~\cref{gradient_function} is the r.h.s. term, which can be easily computed as $-(\mA^{-\frac{1}{2}})^2\frac{\partial l}{\partial \mA^{-\frac{1}{2}}}(\mA^{-\frac{1}{2}})^2$ with $3$ matrix multiplications. For the new iterative solver of the Lyapunov equation $\mB\mX{+}\mX\mB{=}\mC$, we have the following initialization:
\begin{equation}
    \begin{gathered}
    \mB_{0}=\frac{\mA^{-\frac{1}{2}}}{||\mA^{-\frac{1}{2}}||_{\rm F}}=||\mA^{\frac{1}{2}}||_{\rm F}\mA^{-\frac{1}{2}}\\
    \mC_{0}=\frac{-\mA^{-1}\frac{\partial l}{\partial \mA^{-\frac{1}{2}}}\mA^{-1}}{||\mA^{-\frac{1}{2}}||_{\rm F}}=-||\mA^{\frac{1}{2}}||_{\rm F}\mA^{-1}\frac{\partial l}{\partial \mA^{-\frac{1}{2}}}\mA^{-1}.
    \end{gathered}
\end{equation}
Then we use the coupled NS iteration to compute the gradient $\frac{\partial l}{\partial \mA}{=}\frac{1}{2}\mC_{k}$. Table~\ref{tab:backward_complexiy} presents the complexity of the backward algorithms. Compared with the gradient of matrix square root, this extension marginally increases the computational complexity by $3$ more matrix multiplications, which is more efficient than a matrix inverse or solving a linear system.

\section{Experiments}\label{sec:exp}

In the experimental section, we first perform a series of numerical tests to compare our proposed method with SVD and NS iteration. Subsequently, we evaluate our methods in several real-world applications, including decorrelated batch normalization, second-order vision transformer, global covariance pooling for image/video recognition, and neural style transfer. The implementation details are kindly referred to the Supplementary Material.

\subsection{Baselines}

In the numerical tests, we compare our two methods against SVD and NS iteration. For the various computer vision experiments, our methods are compared with more differentiable SVD baselines where each one has  its specific gradient computation. These methods include (1) Power Iteration (PI), (2) SVD-PI~\cite{wang2019backpropagation}, (3) SVD-Taylor~\cite{wang2021robust,song2021approximate}, and (4) SVD-Pad\'e~\cite{song2021approximate}. We put the detailed illustration of baseline methods in the Supplementary Material. 

\subsection{Numerical Tests}

To comprehensively evaluate the numerical performance and stability, we compare the speed and error for the input of different batch sizes, matrices in various dimensions, different iteration times of the backward pass, and different polynomial degrees of the forward pass. In each of the following tests, the comparison is based on $10,000$ random covariance matrices and the matrix size is consistently $64{\times}64$ unless explicitly specified. The error is measured by calculating the Mean Absolute Error (MAE) and Normalized Root Mean Square Error (NRMSE) of the matrix square root computed by the approximate methods (NS iteration, MTP, and MPA) and the accurate method (SVD).

For our algorithm of fast inverse square root, since the theory behind the algorithm is in essence the same with the matrix square root, they are expected to have similar numerical properties. The difference mainly lie in the forward error and backward speed. Thereby, we conduct the FP error analysis and the BP speed analysis for the inverse square root in Sec.~\ref{sec:fp_err_speed} and Sec.~\ref{sec:bp_speed}, respectively. For the error analysis, we compute the error of whitening transform by $||\sigma(\mA^{-\frac{1}{2}}\mX){-}\mI||_{\rm F}$ where $\sigma(\cdot)$ denotes the extracted eigenvalues. In the other numerical tests, we only evaluate the properties of the algorithm for the matrix square root.

\begin{figure}[htbp]
    \centering
    \includegraphics[width=0.99\linewidth]{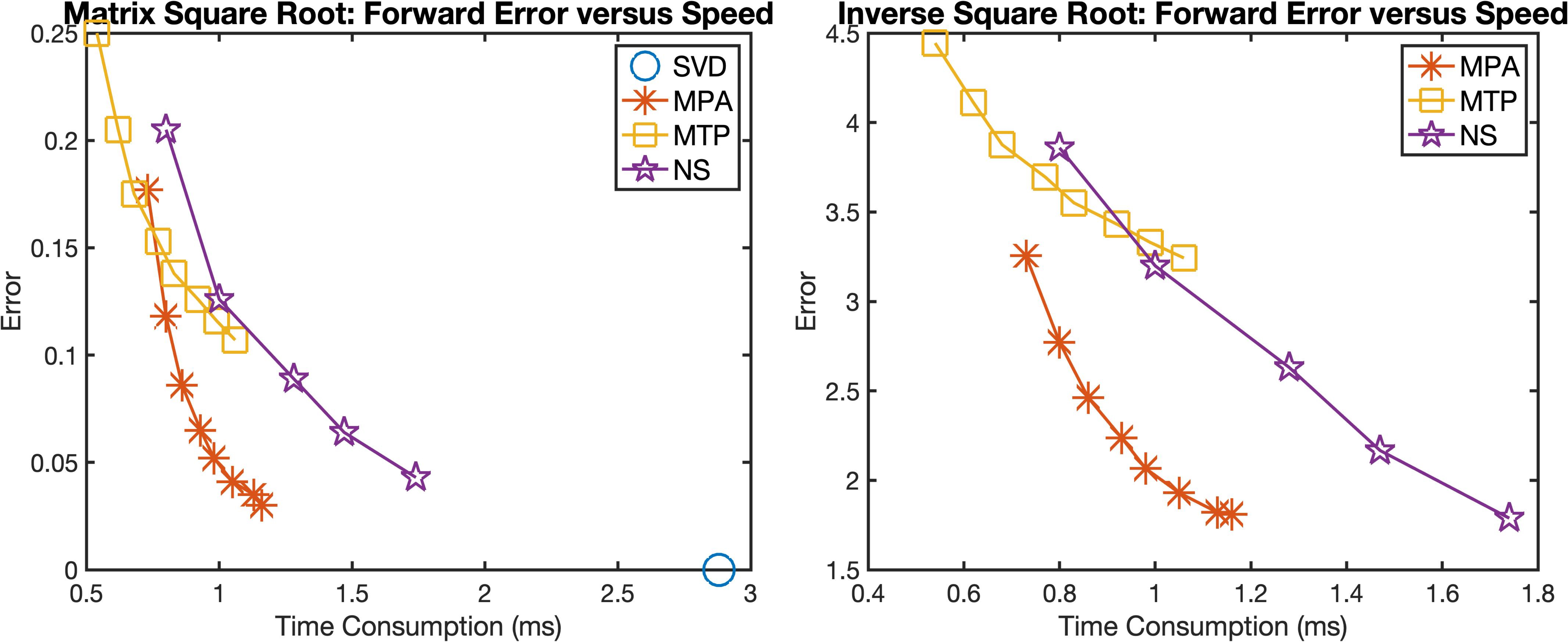}
    \caption{The comparison of speed and error in the FP for the matrix square root (\emph{left}) and the inverse square root (\emph{right}). Our MPA computes the more accurate and faster solution than the NS iteration, and our MTP enjoys the fastest calculation speed. }
    \label{fig:fp_sped_err}
\end{figure}

\subsubsection{Forward Error versus Speed}
\label{sec:fp_err_speed}
Both the NS iteration and our methods have a hyper-parameter to tune in the forward pass, \emph{i.e.,} iteration times for NS iteration and polynomial degrees for our MPA and MTP. To validate the impact, we measure the speed and error of both matrix square root and its inverse for different hyper-parameters. The degrees of our MPA and MTP vary from $6$ to $18$, and the iteration times of NS iteration range from $3$ to $7$. As can be observed from Fig.~\ref{fig:fp_sped_err}, our MTP has the least computational time, and our MPA consumes slightly more time than MTP but provides a closer approximation. Moreover, the curve of our MPA consistently lies below that of the NS iteration, demonstrating our MPA is a better choice in terms of both speed and accuracy.

\begin{figure}[htbp]
    \centering
    \includegraphics[width=0.49\linewidth]{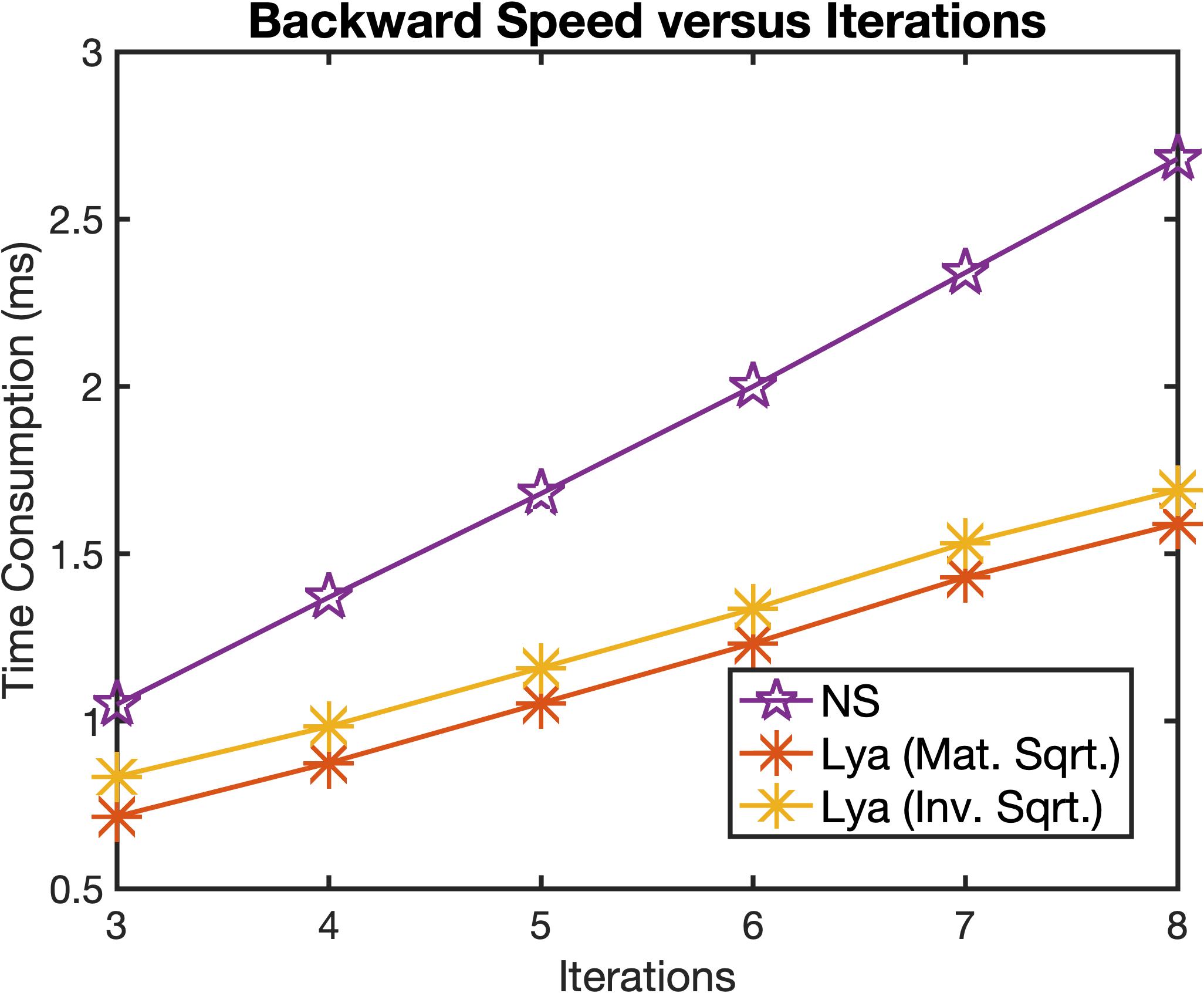}
    \caption{The speed comparison in the backward pass. Our Lyapunov solver is more efficient than NS iteration as fewer matrix multiplications are involved. Our solver for inverse square root only slightly increases the computational cost.}
    \label{fig:bp_speed}
\end{figure}

\subsubsection{Backward Speed versus Iteration}
\label{sec:bp_speed}
Fig.~\ref{fig:bp_speed} compares the speed of our backward Lyapunov solver and the NS iteration versus different iteration times. The result is coherent with the complexity analysis in Table~\ref{tab:backward_complexiy}: our Lyapunov solver is much more efficient than NS iteration. For the NS iteration of $5$ times, our Lyapunov solver still has an advantage even when we iterate $8$ times. Moreover, the extension of our Lyapunov solver for inverse square root only marginally increases the computational cost and is sill much faster than the NS iteration.

\begin{figure}[htbp]
    \centering
    \includegraphics[width=0.99\linewidth]{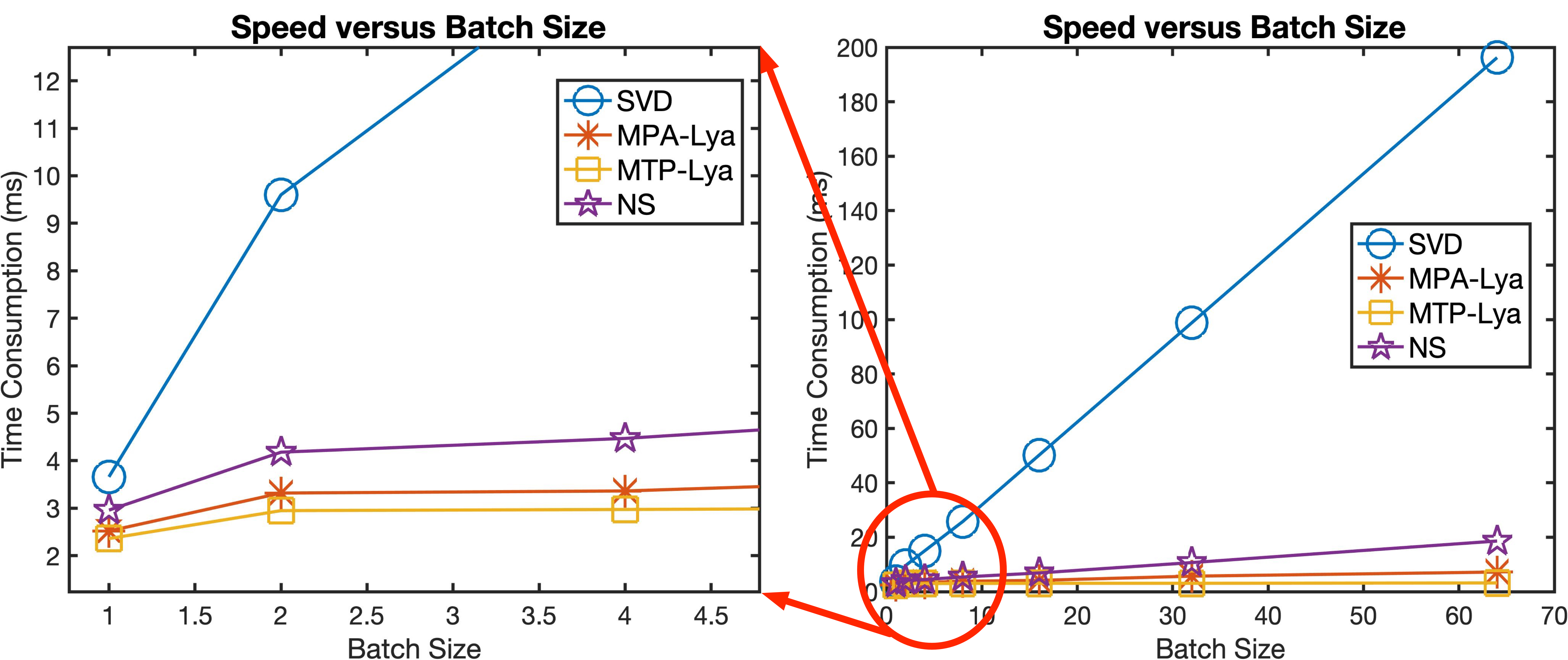}
    \caption{Speed comparison for each method versus different batch sizes. Our methods are more batch-efficient than the SVD or NS iteration. }
    \label{fig:speed_bs}
\end{figure}

\subsubsection{Speed versus Batch Size}
In certain applications such as covariance pooling and instance whitening, the input could be batched matrices instead of a single matrix. To compare the speed for batched input, we conduct another numerical test. The hyper-parameter choices follow our experimental settings in decorrelated batch normalization. As seen in Fig.~\ref{fig:speed_bs}, our MPA-Lya and MTP-Lya are consistently more efficient than the NS iteration and SVD. To give a concrete example, when the batch size is $64$, our MPA-Lya is $2.58$X faster than NS iteration and $27.25$X faster than SVD, while our MTP-Lya is $5.82$X faster than the NS iteration and $61.32$X faster than SVD. 

\begin{table*}[htbp]
    \centering
    \caption{Validation error of ZCA whitening methods. The covariance matrix is of size $1{\times}64{\times}64$. The time consumption is measured for computing the inverse square root (BP+FP). For each method, we report the results based on five runs.}
    \resizebox{0.8\linewidth}{!}{
    \begin{tabular}{r|c|c|c|c|c|c|c}
    \hline 
        \multirow{3}*{Methods} & \multirow{3}*{ Time (ms)} & \multicolumn{4}{c|}{ResNet-18} & \multicolumn{2}{c}{ResNet-50}\\
        \cline{3-8}
        & & \multicolumn{2}{c|}{CIFAR10} & \multicolumn{2}{c|}{CIFAR100} & \multicolumn{2}{c}{CIFAR100}\\
        \cline{3-8}
        && mean$\pm$std & min & mean$\pm$std & min & mean$\pm$std & min \\
        \hline
        SVD-Clip &3.37 & 4.88$\pm$0.25 &4.65 & 21.60$\pm$0.39 &21.19 & 20.50$\pm$0.33 &20.17\\
        SVD-PI (GPU) &5.27 &4.57$\pm$0.10 &4.45 &21.35$\pm$0.25 &21.05 &19.97$\pm$0.41 & 19.27 \\
        SVD-PI & 3.49 & 4.59$\pm$0.09 &4.44 &21.39$\pm$0.23 &21.04 &19.94$\pm$0.44 &19.28 \\
        SVD-Taylor &3.41 &4.50$\pm$0.08 &4.40 &21.14$\pm$0.20 &\textbf{20.91} &19.81$\pm$0.24&19.26\\
        SVD-Pad\'e &3.39 & 4.65$\pm$0.11 &4.50 &21.41$\pm$0.15 &21.26 &20.25$\pm$0.23&19.98\\
        NS Iteration & 2.96 & 4.57$\pm$0.15 &4.37&21.24$\pm$0.20 &21.01&\textbf{19.39$\pm$0.30}&\textbf{19.01}\\
        \hline
        Our MPA-Lya & 2.61 &\textbf{4.39$\pm$0.09} &\textbf{4.25} & \textbf{21.11}$\pm$\textbf{0.12}&20.95& \textbf{19.55$\pm$0.20} &19.24\\
        Our MTP-Lya & \textbf{2.56} & 4.49$\pm$0.13 &4.31 & 21.42$\pm$0.21 &21.24 &20.55$\pm$0.37&20.12\\
     \hline 
    \end{tabular}
   }
    \label{tab:zca_whitening}
\end{table*}

As discussed before, the current SVD implementation adopts a for-loop to compute each matrix one by one within the mini-batch. This accounts for why the time consumption of SVD grows almost linearly with the batch size. For the NS iteration, the backward pass is not as batch-friendly as our Lyapunov solver. The gradient calculation requires measuring the trace and handling the multiplication for each matrix in the batch, which has to be accomplished ineluctably by a for-loop. Our backward pass can be more efficiently implemented by batched matrix multiplication. 

\begin{figure}[htbp]
    \centering
    \includegraphics[width=0.99\linewidth]{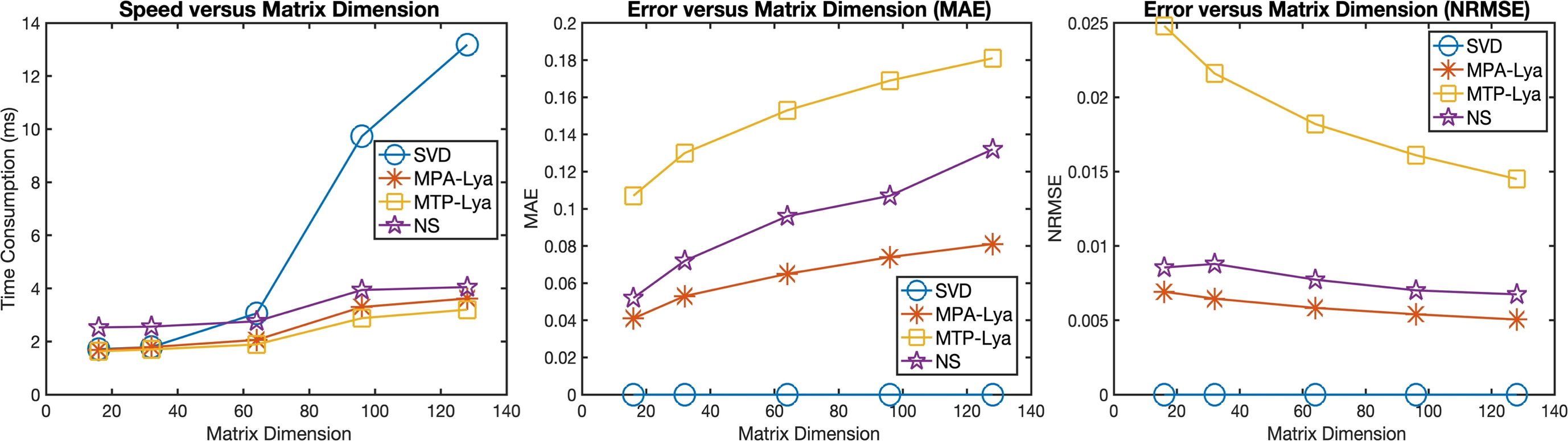}
    \caption{The speed comparison (\emph{left}) and the error comparison (\emph{middle and right}) for matrices in different dimensions. Our MPA-Lya is consistently faster and more accurate than NS iteration for different matrix dimensions. Since the SVD is accurate by default, other approximate methods are compared with SVD to measure the error.}
    \label{fig:speed_err_dim}
\end{figure}

\subsubsection{Speed and Error versus Matrix Dimension}
In the last numerical test, we compare the speed and error for matrices in different dimensions. The hyper-parameter settings also follow our experiments of ZCA whitening. As seen from Fig.~\ref{fig:speed_err_dim} left, our proposed MPA-Lya and MTP-Lya consistently outperform others in terms of speed. In particular, when the matrix size is very small (${<}32$), the NS iteration does not hold a speed advantage over the SVD. By contrast, our proposed methods still have competitive speed against the SVD. Fig.~\ref{fig:speed_err_dim} right presents the approximation error using metrics MAE and NRMSE. Both metrics agree well with each other and demonstrate that our MPA-Lya always has a better approximation than the NS iteration, whereas our MTP-Lya gives a worse estimation but takes the least time consumption, which can be considered as a trade-off between speed and accuracy. 

\subsection{Decorrelated Batch Normalization}

As a substitute of ordinary BN, the decorrelated BN~\cite{huang2018decorrelated} applies the ZCA whitening transform to eliminate the correlation of the data. Consider the reshaped feature map $\mX{\in}\mathbb{R}^{C{\times} BHW}$. The whitening procedure first computes its sample covariance as:
\begin{equation}
    \mA{=}(\mX-\mu(\mX))(\mX-\mu(\mX))^{T}{+}\epsilon\mI
    \label{zca_cov}
\end{equation}
where $\mA{\in}\mathbb{R}^{C{\times}C}$, $\mu(\mX)$ is the mean of $\mX$, and $\epsilon$ is a small constant to make the covariance strictly positive definite. Afterwards, the inverse square root is calculated to whiten the feature map:
\begin{equation}
    \mX_{whitend}=\mA^{-\frac{1}{2}}\mX
\end{equation}
By doing so, the eigenvalues of $\mX$ are all ones, \emph{i.e.,} the feature is uncorrelated. During the training process, the training statistics are stored for the inference phase. We insert the decorrelated BN layer after the first convolutional layer of ResNet~\cite{he2016deep}, and the proposed methods and other baselines are used to compute $\mA^{-\frac{1}{2}}$.

Table~\ref{tab:zca_whitening} displays the speed and validation error on CIFAR10 and CIFAR100~\cite{krizhevsky2009learning}. The ordinary SVD with clipping gradient (SVD-Clip) is inferior to other SVD baselines, and the SVD computation on GPU is slower than that on CPU. Our MTP-Lya is $1.16$X faster than NS iteration and $1.32$X faster than SVD-Pad\'e, and our MPA-Lya is $1.14$X and $1.30$X faster. Furthermore, our MPA-Lya achieves state-of-the-art performances across datasets and models. Our MTP-Lya has comparable performances on ResNet-18 but slightly falls behind on ResNet-50. We guess this is mainly because the relatively large approximation error of MTP might affect little on the small model but can hurt the large model. On CIFAR100 with ResNet-50, our MPA-Lya slightly falls behind NS iteration in the average validation error. 
As a larger and deeper model, ResNet-50 is likely to have worse-conditioned matrices than ResNet-18. Since our MPA involves solving a linear system, processing a very ill-conditioned matrix could lead to some round-off errors. In this case, NS iteration might have a chance to slightly outperform our MPA-Lya. However, this is a rare situation; our MPA-Lya beats NS iteration in most following experiments.

\subsection{Global Covariance Pooling}

For the application of global covariance pooling, we evaluate our method in three different tasks, including large-scale visual recognition, fine-grained visual categorization, and video action recognition. Since the GCP method requires the very accurate matrix square root~\cite{song2021approximate}, our MTP-Lya cannot achieve reasonable performances due to the relatively large approximation error. Therefore, we do not take it into account for comparison throughout the GCP experiments.

\subsubsection{Large-scale Visual Recognition}

\begin{figure}[htbp]
    \centering
    \includegraphics[width=0.99\linewidth]{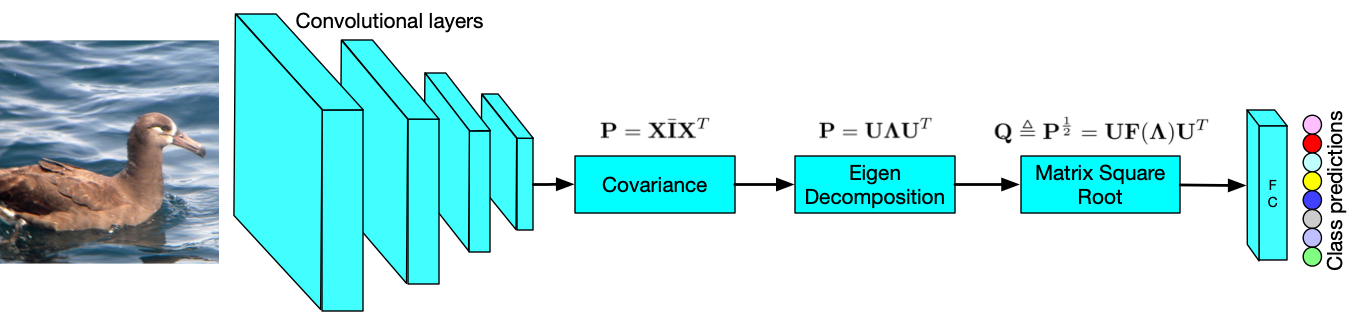}
    \caption{Overview of the GCP network~\cite{li2017second,li2018towards,song2021approximate} for large-scale and fine-grained visual recognition.}
    \label{fig:arch_gcp}
\end{figure}

Fig.~\ref{fig:arch_gcp} displays the architecture of a typical GCP network. Different from the standard CNNs, the covariance square root of the last convolutional feature is used as the global representation. Considering the final convolutional feature $\mX{\in}\mathbb{R}^{B{\times}C{\times}HW}$, a GCP meta-layer first computes the sample covariance as:
\begin{equation}
    \mathbf{P}=\mathbf{X}\Bar{\mathbf{I}}\mathbf{X}^{T},\ \Bar{\mathbf{I}}=\frac{1}{N}(\mathbf{I}-\frac{1}{N}\mathbf{1}\mathbf{1}^{T})
    \label{covariance}
\end{equation}
where $\Bar{\mathbf{I}}$ represents the centering matrix, $\mathbf{I}$ denotes the identity matrix, and $\mathbf{1}$ is a column vector whose values are all ones, respectively. Afterwards, the matrix square root is conducted for normalization:
\begin{equation}
    \mathbf{Q}\triangleq\mathbf{P}^{\frac{1}{2}}=(\mathbf{U}\mathbf{\Lambda}\mathbf{U}^{T})^{\frac{1}{2}}=\mU\mathbf{\Lambda}^{\frac{1}{2}}\mathbf{U}^{T}
    \label{matrix_power}
\end{equation}
where the normalized covariance matrix $\mathbf{Q}$ is fed to the FC layer. Our method is applied to calculate $\mQ$.

\begin{table}[htbp]
\caption{Comparison of validation accuracy (\%) on ImageNet~\cite{deng2009imagenet} and ResNet-50~\cite{he2016deep}. The covariance is of size {$256{\times}256{\times}256$}, and the time consumption is measured for computing the matrix square root (FP+BP).}
    \centering
    \resizebox{0.99\linewidth}{!}{
    \begin{tabular}{r|c|c|c}
    \hline
        Methods & Time (ms)& Top-1 Acc. & Top-5 Acc.  \\
        \hline
        SVD-Taylor &2349.12 &77.09 &93.33 \\
        SVD-Pad\'e &2335.56 &\textbf{77.33} &\textbf{93.49} \\
        NS iteration  &164.43 & 77.19 & 93.40\\
        \hline
        Our MPA-Lya & \textbf{110.61} &77.13 & 93.45\\
    \hline
    \end{tabular}
    }
    \label{tab:performances_GCP_CNN}
\end{table}

Table~\ref{tab:performances_GCP_CNN} presents the speed comparison and the validation error of GCP ResNet-50~\cite{he2016deep} models on ImageNet~\cite{deng2009imagenet}. Our MPA-Lya not only achieves very competitive performance but also has the least time consumption. The speed of our method is about $21$X faster than the SVD and $1.5$X faster than the NS iteration.



\subsubsection{Fine-grained Visual Recognition }

\begin{table}[htbp]
\caption{Comparison of validation accuracy on fine-grained benchmarks and ResNet-50~\cite{he2016deep}. The covariance is of size {$10{\times}64{\times}64$}, and the time consumption is measured for computing the matrix square root (FP+BP).}
    \centering
    \resizebox{0.99\linewidth}{!}{
    \begin{tabular}{r|c|c|c|c}
    \hline
        Methods & Time (ms)& Birds & Aircrafts & Cars  \\
        \hline
        SVD-Taylor &32.13 &86.9 &89.9 &92.3 \\
        SVD-Pad\'e &31.54 &87.2 &90.5 &\textbf{92.8} \\
        NS iteration  &5.79 & 87.3 & 89.5 & 91.7 \\
        \hline
        Our MPA-Lya & \textbf{3.89} &\textbf{87.8} &\textbf{91.0} &92.5 \\
    \hline
    \end{tabular}
    }
    \label{tab:performances_GCP_fgvc}
\end{table}

In line with other GCP works~\cite{li2017second,li2018towards,song2021approximate}, after training on ImageNet, the model is subsequently fine-tuned on each fine-grained dataset. Table~\ref{tab:performances_GCP_fgvc} compares the time consumption and validation accuracy on three commonly used fine-grained benchmarks, namely Caltech University Birds (Birds)~\cite{WelinderEtal2010}, FGVC Aircrafts (Aircrafts)~\cite{maji2013fine}, and Stanford Cars (Cars)~\cite{KrauseStarkDengFei-Fei_3DRR2013}. As can be observed, our MPA-Lya consumes $50\%$ less time than the NS iteration and is about $8$X faster than the SVD. Moreover, the performance of our method is slightly better than other baselines on Birds~\cite{WelinderEtal2010} and Aircrafts~\cite{maji2013fine}. The evaluation result on Cars~\cite{KrauseStarkDengFei-Fei_3DRR2013} is also comparable.

\subsubsection{Video Action Recognition}

\begin{figure}[htbp]
    \centering
    \includegraphics[width=0.99\linewidth]{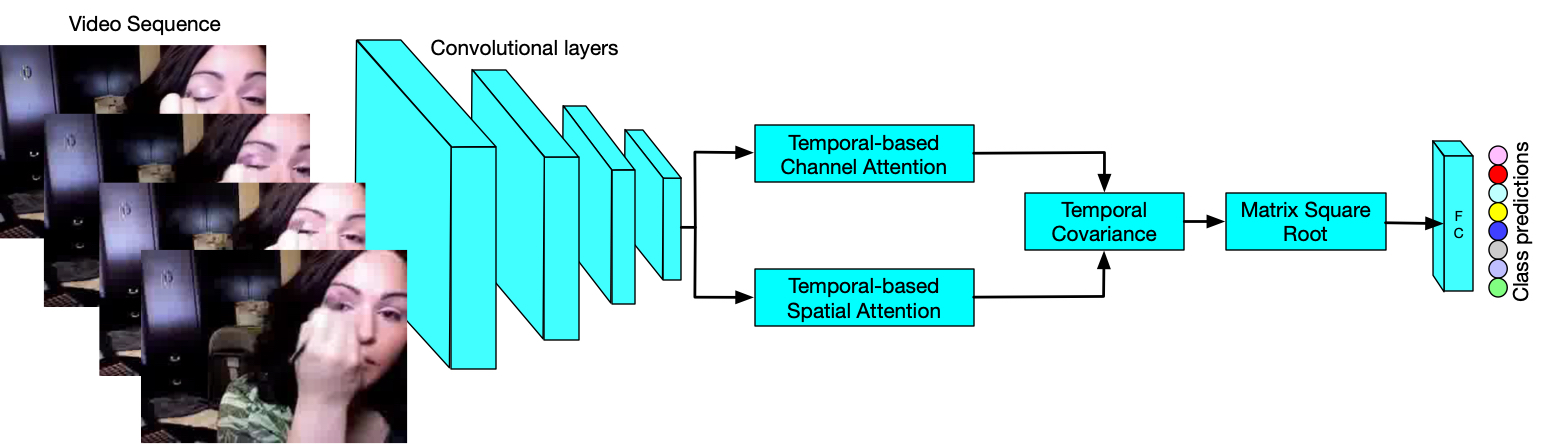}
    \caption{Architecture of the temporal-attentive GCP network for video action recognition~\cite{gao2021temporal}. The channel and spatial attention is used to make the covariance more attentive.}
    \label{fig:arch_gcp_video}
\end{figure}

Besides the application of image recognition, the GCP methods can be also used for the task of video recognition~\cite{gao2021temporal}. Fig.~\ref{fig:arch_gcp_video} displays the overview of the temporal-attentive GCP model for video action recognition. The temporal covariance is computed in a sliding window manner by involving both intra- and inter-frame correlations. Supposing the kernel size of the sliding window is $3$, then temporal covariance is computed as:
\begin{equation}
\begin{gathered}
     Temp.Cov.(\mathbf{X}_{l})=\underbrace{\mX_{l-1}\mX_{l-1}^{T} + \mX_{l}\mX_{l}^{T} + \mX_{l+1}\mX_{l+1}^{T}}_{intra-frame\ covariance}\\
     +\underbrace{\mX_{l-1}\mX_{l}^{T} + \mX_{l}\mX_{l-1}^{T} + \cdots + \mX_{l+1}\mX_{l}^{T}}_{inter-frame\ covariance} 
\end{gathered}
\end{equation}
Finally, the matrix square root of the attentive temporal-based covariance is computed and passed to the FC layer. The spectral methods are used to compute the matrix square root of the attentive covariance $Temp.Cov.(\mathbf{X}_{l})$. 

\begin{table}[htbp]
    \centering
    \caption{Validation top-1/top-5 accuracy (\%) on HMBD51~\cite{Kuehne11} and UCF101~\cite{soomro2012ucf101} with backbone TEA R50~\cite{li2020tea}. The covariance matrix is of size $16{\times}128{\times}128$, and the time consumption is measured for computing the matrix square root (BP+FP).}
    \resizebox{0.99\linewidth}{!}{
    \begin{tabular}{r|c|c|c}
    \hline 
        Methods & Time (ms) & HMBD51 & UCF101 \\
        \hline
        SVD-Taylor &76.17 &73.79/93.84 &\textbf{95.00}/\textbf{99.60}   \\
        SVD-Pad\'e &75.25 &73.89/93.79 &94.13/99.47 \\
        NS Iteration &12.11  &72.75/93.86 &94.16/99.50 \\
        \hline
        Our MPA-Lya  &\textbf{6.95} &\textbf{74.05}/\textbf{93.99} &94.24/99.58 \\
     \hline 
    \end{tabular}
    }
    \label{tab:video_gcp}
\end{table}

We present the validation accuracy and time cost for the video action recognition in Table~\ref{tab:video_gcp}. For the computation speed, our MPA-Lya is about $1.74$X faster than the NS iteration and is about $10.82$X faster than the SVD. Furthermore, our MPA-Lya achieves the best performance on HMDB51, while the result on UCF101 is also very competitive.

To sum up, our MPA-Lya has demonstrated its general applicability in the GCP models for different tasks. In particular, without the sacrifice of performance, our method can bring considerable speed improvements. This could be beneficial for faster training and inference. In certain experiments such as fine-grained classification, the approximate methods (MPA-Lya and NS iteration) can marginally outperform accurate SVD. This phenomenon has been similarly observed in related studies~\cite{li2018towards,huang2019iterative,song2021approximate}, and one likely reason is that 
the SVD does not have as healthy gradients as the approximate methods. This might negatively influence the optimization process and consequently the performance would degrade. 

\subsection{Neural Style Transfer}

\begin{figure}[htbp]
    \centering
    \includegraphics[width=0.9\linewidth]{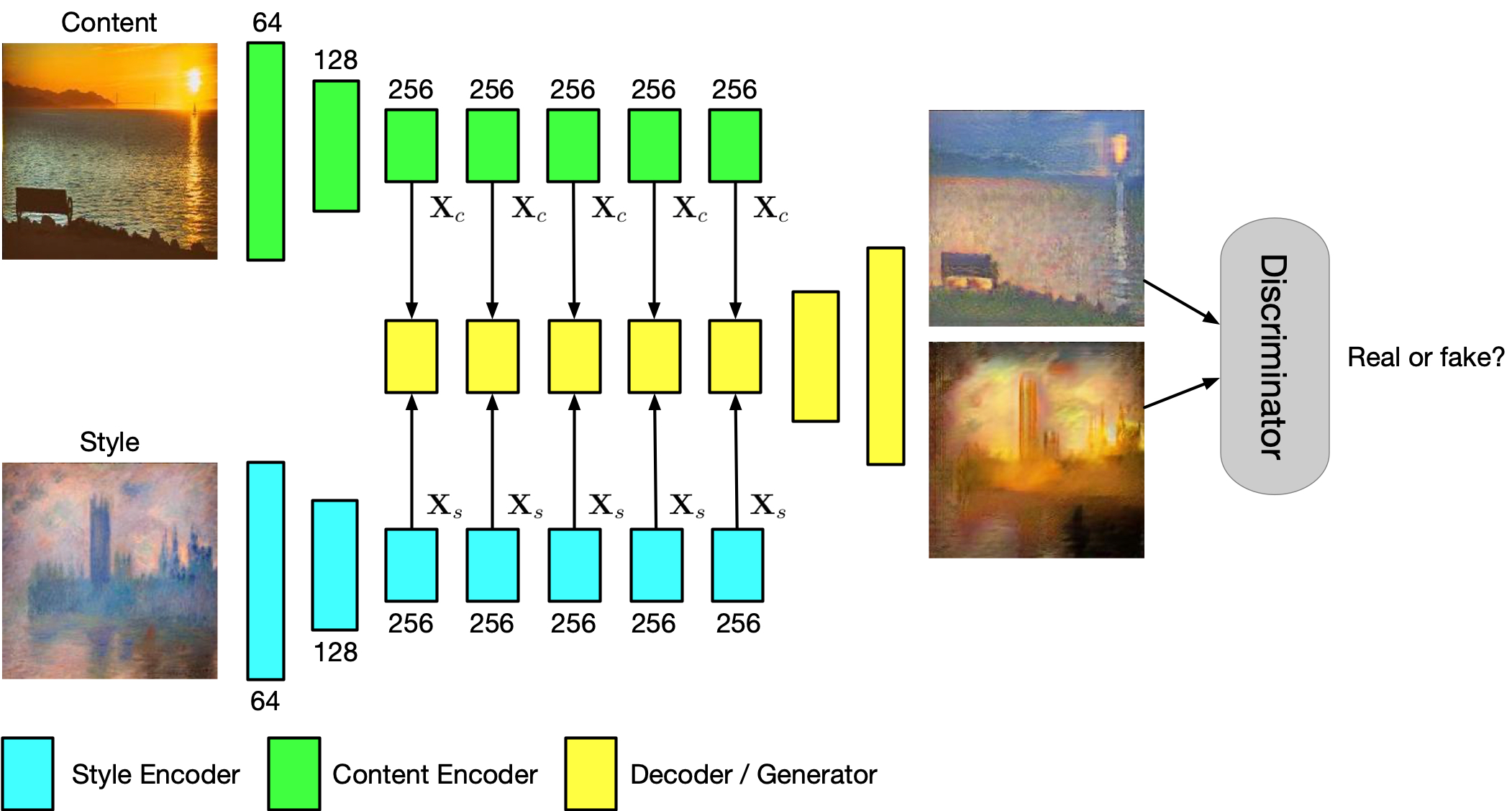}
    \caption{The architecture overview of our model for neural style transfer. Two encoders take input of the style and content image respectively, and generate the multi-scale content/style features. A decoder is applied to absorb the feature and perform the WCT process at $5$ different scales, which outputs a pair of images that exchange the styles. Finally, a discriminator is further adopted to tell apart the authenticity of the images.}
    \label{fig:arch_style_transfer}
\end{figure}

We adopt the WCT process in the network architecture proposed in Cho~\emph{et al.}~\cite{cho2019image} for neural style transfer. Fig.~\ref{fig:arch_style_transfer} displays the overview of the model. The WCT performs successive whitening and coloring transform on the content and style feature. Consider the reshaped content feature $\mathbf{X}_{c}{\in}\mathrm{R}^{B{\times}C{\times}HW}$ and the style feature $\mathbf{X}_{s}{\in}\mathrm{R}^{B{\times}C{\times}HW}$. The style information is first removed from the content as:
\begin{equation}
    \begin{gathered}
    \mathbf{X}_{c}^{whitened} = \Big((\mathbf{X}_{c}-\mu(\mathbf{X}_{c}))(\mathbf{X}_{c}-\mu(\mathbf{X}_{c}))^{T}\Big)^{-\frac{1}{2}}\mathbf{X}_{c}
    \end{gathered}
\end{equation}
Then we extract the desired style information from the style feature $\mathbf{X}_{s}$ and transfer it to the whitened content feature:
\begin{equation}
    \mathbf{X}_{c}^{colored} = \Big((\mathbf{X}_{s}-\mu(\mathbf{X}_{s}))(\mathbf{X}_{s}-\mu(\mathbf{X}_{s}))^{T}\Big)^{\frac{1}{2}}\mathbf{X}_{c}^{whitened}
\end{equation}
The resultant feature $\mathbf{X}_{c}^{colored}$ is compensated with the mean of style feature and combined with the original content feature:
\begin{equation}
    \mathbf{X} = \alpha (\mathbf{X}_{c}^{colored}+\mu(\mathbf{X}_{s})) + (1-\alpha)\mathbf{X}_{c} 
\end{equation}
where $\alpha$ is a weight bounded in $[0,1]$ to control the strength of style transfer. In this experiment, both the matrix square root and inverse square root are computed.

\begin{table}[htbp]
\caption{The LPIPS~\cite{zhang2018perceptual} score and user preference (\%) on Artworks~\cite{isola2017image} dataset. The covariance is of size $4{\times}256{\times}256$. We measure the time consumption of whitening and coloring transform that is conducted $10$ times to exchange the style and content feature at different network depths.}
    \centering
    \setlength{\tabcolsep}{1.5pt}
    \resizebox{0.99\linewidth}{!}{
    \begin{tabular}{r|c|c|c}
    \hline
        Methods & Time (ms) & LPIPS~\cite{zhang2018perceptual} ($\uparrow$) & Preference ($\uparrow$) \\
        \hline
        SVD-Taylor &447.12 & 0.5276 & 16.25\\
        SVD-Pad\'e &445.23 & 0.5422 & 19.25\\
        NS iteration &94.37 & 0.5578 & 17.00\\
        \hline
        Our MPA-Lya &69.23 &\textbf{0.5615} & \textbf{24.75}\\
        Our MTP-Lya &\textbf{40.97} &0.5489 & 18.50\\
    \hline
    \end{tabular}
    }
    \label{tab:style_transfer_sum}
\end{table}

\begin{figure}[htbp]
    \centering
    \includegraphics[width=0.9\linewidth]{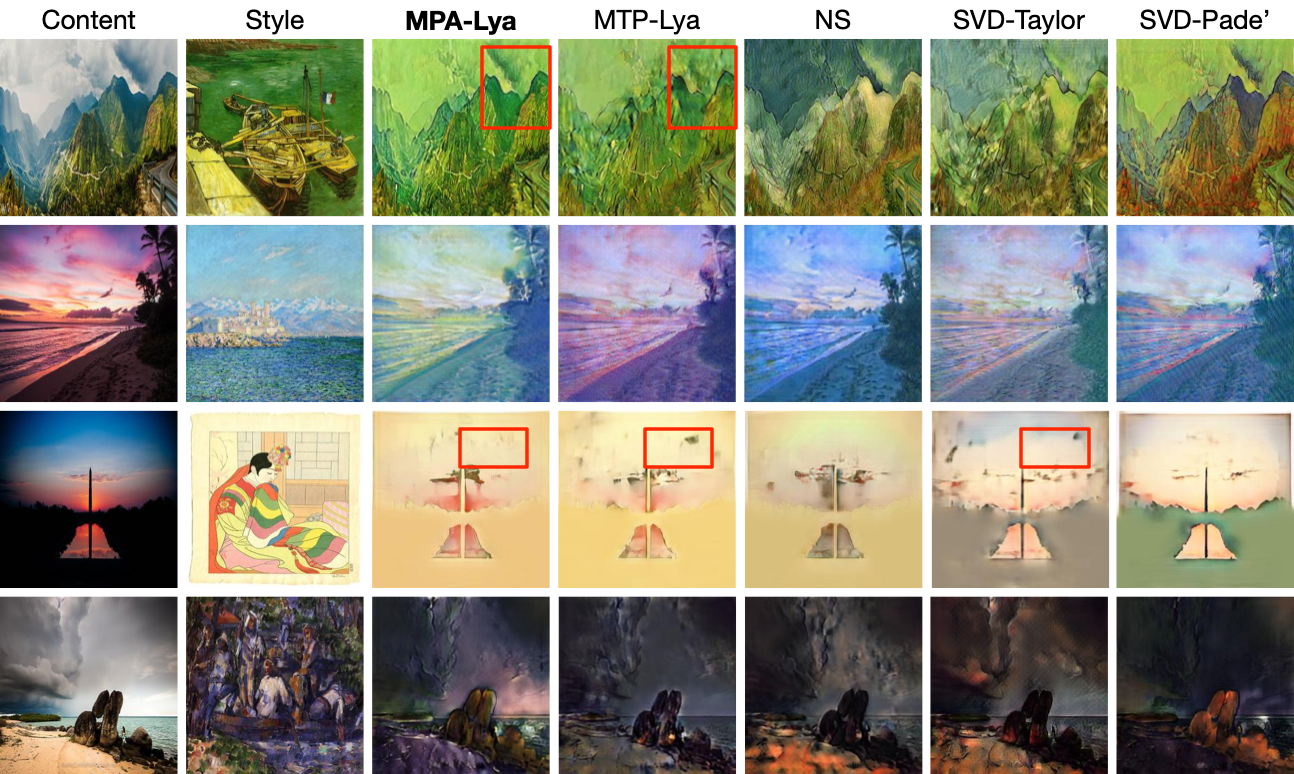}
    \caption{Visual examples of the neural style transfer on Artworks~\cite{isola2017image} dataset. Our methods generate sharper images with more coherent style and better visual appeal. The red rectangular indicates regions with subtle details.}
    \label{fig:style_transfer_visual}
\end{figure}

Table~\ref{tab:style_transfer_sum} presents the quantitative evaluation using the LPIPS~\cite{zhang2018perceptual} score and user preference. The speed of our MPA-Lya and MTP-Lya is significantly faster than other methods. Specifically, our MTP-Lya is $2.3$X faster than the NS iteration and $10.9$X faster than the SVD, while our MPA-Lya consumes $1.4$X less time than the NS iteration and $6.4$X less time than the SVD. Moreover, our MPA-Lya achieves the best LPIPS score and user preference. The performance of our MTP-Lya is also very competitive. Fig.~\ref{fig:style_transfer_visual} displays the exemplary visual comparison. Our methods can effectively transfer the style information and preserve the original content, leading to transferred images with a more coherent style and better visual appeal. 
We give detailed evaluation results on each subset and more visual examples in Supplementary Material.

\begin{table*}[htbp]
    \centering
    \caption{Validation top-1/top-5 accuracy of the second-order vision transformer on ImageNet~\cite{deng2009imagenet}. The covariance is of size $64{\times}48{\times}48$, where $64$ is the mini-batch size. The time cost is measured for computing the matrix square root (BP+FP).} 
     \resizebox{0.79\linewidth}{!}{
    \begin{tabular}{r|c|c|c|c}
    \hline
        \multirow{2}*{Methods} & \multirow{2}*{ Time (ms)} & \multicolumn{3}{c}{Architecture} \\ 
        \cline{3-5}
        & &  So-ViT-7 & So-ViT-10  & So-ViT-14 \\
        \hline
        PI & \textbf{1.84}  & 75.93/93.04 & 77.96/94.18  & 82.16/96.02 (303 epoch)\\
        SVD-PI & 83.43 & 76.55/93.42 & 78.53/94.40 & 82.16/96.01 (278 epoch)\\
        SVD-Taylor & 83.29  & 76.66/\textbf{93.52} & 78.64/94.49 & 82.15/96.02 (271 epoch)\\
        SVD-Pad\'e & 83.25  & 76.71/93.49 & 78.77/94.51 & 82.17/96.02 (265 epoch)\\
        NS Iteration & 10.38  & 76.50/93.44  & 78.50/94.44 & 82.16/96.01 (280 epoch)\\
        \hline
        Our MPA-Lya & 3.25 & \textbf{76.84}/93.46 & \textbf{78.83}/\textbf{94.58} & 82.17/96.03 (\textbf{254} epoch)\\
        Our MTP-Lya & 2.39 & 76.46/93.26 & 78.44/94.33 & 82.16/96.02 (279 epoch)\\
    \hline
    \end{tabular}
    }
    \label{tab:vit_imagenet}
\end{table*}

\subsection{Second-order Vision Transformer}

\begin{figure}
    \centering
    \includegraphics[width=0.9\linewidth]{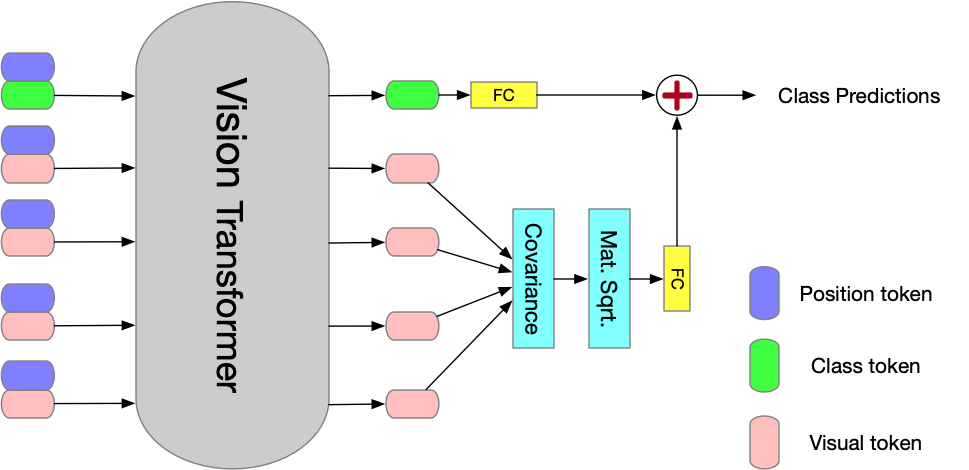}
    \caption{The scheme of So-ViT~\cite{xie2021so}. The covariance square root of the visual tokens are computed to assist the classification. In the original vision transformer~\cite{dosovitskiy2020image}, only the class token is utilized for class predictions.}
    \label{fig:arch_sovit}
\end{figure}

The ordinary vision transformer~\cite{dosovitskiy2020image} attaches an empty class token to the sequence of visual tokens and only uses the class token for prediction, which may not exploit the rich semantics embedded in the visual tokens. Instead, The Second-order Vision Transformer (So-ViT)~\cite{xie2021so} proposes to leverage the high-level visual tokens to assist the task of classification:
\begin{equation}
     y = {\rm FC}(c) + {\rm FC}\Big((\mX\mX^{T})^{\frac{1}{2}}\Big)
\end{equation}
where $c$ is the output class token, $\mX$ denotes the visual token, and $y$ is the combined class predictions. We show the model overview in Fig.~\ref{fig:arch_sovit}. Equipped with the covariance pooling layer, So-ViT removes the need for pre-training on the ultra-large-scale datasets and achieves competitive performance even when trained from scratch. To reduce the computational budget, So-ViT further proposes to use Power Iteration (PI) to approximate the dominant eigenvector. We use our methods to compute the matrix square root of the covariance $\mX\mX^{T}$.

Table~\ref{tab:vit_imagenet} compares the speed and performances on three So-ViT architectures with different depths. Our proposed methods significantly outperform the SVD and NS iteration in terms of speed. To be more specific, our MPA-Lya is $3.19$X faster than the NS iteration and $25.63$X faster than SVD-Pad\'e, and our MTP-Lya is $4.34$X faster than the NS iteration and $34.85$X faster than SVD-Pad\'e. For the So-ViT-7 and So-ViT-10, our MPA-Lya achieves the best evaluation results and even slightly outperforms the SVD-based methods. Moreover, on the So-ViT-14 model where the performances are saturated, our method converges faster and spends fewer training epochs. The performance of our MTP-Lya is also on par with the other methods. The PI suggested in the So-ViT only computes the dominant eigenpair but neglects the rest. In spite of the fast speed, the performance is not comparable with other methods.

\subsection{Ablation Studies}

We conduct three ablation studies to illustrate the impact of the degree of power series in the forward pass, the termination criterion during the back-propagation, and the possibility of combining our Lyapunov solver with the SVD and the NS iteration.  

\subsubsection{Degree of Power series to Match for Forward Pass}

Table~\ref{tab:forward_degree} displays the performance of our MPA-Lya for different degrees of power series. As we use more terms of the power series, the approximation error gets smaller and the performance gets steady improvements from the degree $[3,3]$ to $[5,5]$. When the degree of our MPA is increased from $[5,5]$ to $[6,6]$, there are only marginal improvements. We hence set the forward degrees as $[5,5]$ for our MPA and as $11$ for our MTP as a trade-off between speed and accuracy. 

\begin{table}[htbp]
    \centering
    \setlength{\tabcolsep}{1.5pt}
    \caption{Performance of our MPA-Lya versus different degrees of power series to match.}
   \resizebox{0.99\linewidth}{!}{
    \begin{tabular}{r|c|c|c|c|c|c|c}
    \hline 
        \multirow{3}*{Degrees} & \multirow{3}*{ Time (ms)} & \multicolumn{4}{c|}{ResNet-18} & \multicolumn{2}{c}{ResNet-50}\\
        \cline{3-8}
        & & \multicolumn{2}{c|}{CIFAR10} & \multicolumn{2}{c|}{CIFAR100} & \multicolumn{2}{c}{CIFAR100}\\
        \cline{3-8}
        && mean$\pm$std & min & mean$\pm$std & min & mean$\pm$std & min \\
        \hline
        $[3,3]$ &0.80 &4.64$\pm$0.11&4.54 &21.35$\pm$0.18&21.20  &20.14$\pm$0.43 & 19.56\\
        $[4,4]$ &0.86 &4.55$\pm$0.08&4.51 &21.26$\pm$0.22&21.03  &19.87$\pm$0.29 & 19.64\\
        $[6,6]$ &0.98 &\textbf{4.45$\pm$0.07}&4.33 &\textbf{21.09$\pm$0.14}&21.04  &\textbf{19.51$\pm$0.24}&19.26\\
        \hline
        $[5,5]$ &0.93  &\textbf{4.39$\pm$0.09} &\textbf{4.25} & \textbf{21.11$\pm$0.12} &\textbf{20.95} & \textbf{19.55$\pm$0.20} & \textbf{19.24} \\
     \hline 
    \end{tabular}
    }
    \label{tab:forward_degree}
\end{table}

\subsubsection{Termination Criterion for Backward Pass}
\label{sec:gradient_error}

\begin{table*}[htbp]
    \centering
    \setlength{\tabcolsep}{1.5pt}
    \caption{Performance of our MPA-Lya versus different iteration times. The residual errors $||\mB_{k}{-}\mI||$ and $||0.5\mC_{k}-\mX||_{\rm F}$ are measured based on $10,000$ randomly sampled matrices.}
    \resizebox{0.8\linewidth}{!}{
    \begin{tabular}{r|c|c|c|c|c|c|c|c|c}
    \hline 
        \multirow{3}*{Methods} & \multirow{3}*{ Time (ms)} & \multirow{3}*{$||\mB_{k}{-}\mI||_{\rm F}$} & \multirow{3}*{$||0.5\mC_{k}{-}\mX||_{\rm F}$} & \multicolumn{4}{c|}{ResNet-18} & \multicolumn{2}{c}{ResNet-50}\\
        \cline{5-10}
        & & & & \multicolumn{2}{c|}{CIFAR10} & \multicolumn{2}{c|}{CIFAR100} & \multicolumn{2}{c}{CIFAR100}\\
        \cline{5-10}
        & & & & mean$\pm$std & min & mean$\pm$std & min & mean$\pm$std & min \\
        \hline
        BS algorithm &2.34 &-- &-- & 4.57$\pm$0.10&4.45 &21.20$\pm$0.23&21.01 &\textbf{19.60$\pm$0.16}&19.55 \\
         \#iter 5 &1.14& ${\approx}0.3541$ &${\approx}0.2049$ & 4.48$\pm$0.13&4.31 & 21.15$\pm$0.24&\textbf{20.84} &20.03$\pm$0.19&19.78 \\
         \#iter 6 &1.33& ${\approx}0.0410$ &${\approx}0.0231$  & 4.43$\pm$0.10 &4.28 & 21.16$\pm$0.19 &20.93 & 19.83$\pm$0.24 &19.57 \\
         \#iter 7 &1.52& ${\approx}7e{-}4$ &${\approx}3.5e{-}4$ & 4.45$\pm$0.11&4.29 & 21.18$\pm$0.20&20.95 &19.69$\pm$0.20&19.38 \\
         \#iter 9 &1.83& ${\approx}2e{-}7$ &${\approx}7e{-}6$ & \textbf{4.40$\pm$0.07} &4.28 & \textbf{21.08$\pm$0.15} &20.89 & \textbf{19.52$\pm$0.22} &19.25 \\
        \hline
         \#iter 8 &1.62& ${\approx}3e{-}7$ & ${\approx}7e{-}6$& \textbf{4.39$\pm$0.09}&\textbf{4.25} & \textbf{21.11$\pm$0.12}&20.95 & \textbf{19.55$\pm$0.20}&\textbf{19.24} \\
     \hline 
    \end{tabular}
    }
    \label{tab:back_iteration}
\end{table*}

Table~\ref{tab:back_iteration} compares the performance of backward algorithms with different termination criteria as well as the exact solution computed by the Bartels-Steward algorithm (BS algorithm)~\cite{bartels1972solution}. Since the NS iteration has the property of quadratic convergence, the errors $||\mB_{k}{-}\mI||_{\rm F}$ and $||0.5\mC_{k}-\mX||_{\rm F}$ decrease at a larger rate for more iteration times. When we iterate more than $7$ times, the error becomes sufficiently neglectable, \emph{i.e.,} the NS iteration almost converges. Moreover, from $8$ iterations to $9$ iterations, there are no obvious performance improvements. We thus terminate the iterations after iterating $8$ times. 


The exact gradient calculated by the BS algorithm does not yield the best results. Instead, it only achieves the least fluctuation on ResNet-50 and other results are inferior to our iterative solver. This is because the formulation of our Lyapunov equation is based on the assumption that the accurate matrix square root is computed, but in practice we only compute the approximate one in the forward pass. In this case, calculating \textit{the accurate gradient of the approximate matrix square root} might not necessarily work better than \textit{the approximate gradient of the approximate matrix square root}. 


\subsubsection{Lyapunov Solver as A General Backward Algorithm}
\label{sec:lya_backward}


We note that our proposed iterative Lyapunov solver is a general backward algorithm for computing the matrix square root. That is to say, it should be also compatible with the SVD and NS iteration as the forward pass. 

For the NS-Lya, our previous conference paper~\cite{song2022fast} shows that the NS iteration used in~\cite{higham2008functions,li2017second} cannot converge on any datasets. In this extended manuscript, we found out that the underlying reason is the inconsistency between the FP and BP. The NS iteration of~\cite{higham2008functions,li2017second} is a coupled iteration that use two variables $\mY_{k}$ and $\mZ_{k}$ to compute the matrix square root. For the BP algorithm, the NS iteration is defined to compute the matrix sign and only uses one variable $\mY_{k}$. The term $\mZ_{k}$ is not involved in the BP and we have no control over the gradient back-propagating through it, which results in the non-convergence of the model. To resolve this issue, we propose to change the forward coupled NS iteration to a variant that uses one variable as:
\begin{equation}
    \mZ_{k+1}=\frac{1}{2}(3\mZ_{k}-\mZ_{k}^{3}\frac{\mA}{||\mA||_{\rm F}})
\end{equation}
where $\mZ_{k+1}$ converges to the inverse square root $\mA^{-\frac{1}{2}}$. This variant of NS iteration is often used to directly compute the inverse square root~\cite{huang2019iterative,bini2005algorithms}. 
The $\mZ_{0}$ is initialization with $\mI$, and post-compensation is calculated as $\mZ_{k}=\frac{1}{\sqrt{||\mA||_{\rm F}}} \mZ_{k}$. Although the modified NS iteration uses only one variable, we note that it is an equivalent representation with the previous NS iteration. More formally, we have:
\begin{prop}
 The one-variable NS iteration of~\cite{huang2019iterative,bini2005algorithms} is equivalent to the two-variable NS iteration of~\cite{li2017second,lin2017improved,higham2008functions}. 
\end{prop}

We give the proof in the Supplementary Material. The modified forward NS iteration is compatible with our iterative Lyapunov solver. Table~\ref{tab:abla_combination} compares the performance of different methods that use the Lyapunov solver as the backward algorithm. Both the SVD-Lya and NS-Lya achieve competitive performances. 


\begin{table}[htbp]
    \centering
    \setlength{\tabcolsep}{1.5pt}
    \caption{Performance comparison of SVD-Lya and NS-Lya.}
    \resizebox{0.99\linewidth}{!}{
    \begin{tabular}{r|c|c|c|c|c|c|c}
    \hline 
        \multirow{3}*{Methods} & \multirow{3}*{ Time (ms)} & \multicolumn{4}{c|}{ResNet-18} & \multicolumn{2}{c}{ResNet-50}\\
        \cline{3-8}
        & & \multicolumn{2}{c|}{CIFAR10} & \multicolumn{2}{c|}{CIFAR100} & \multicolumn{2}{c}{CIFAR100}\\
        \cline{3-8}
        && mean$\pm$std & min & mean$\pm$std & min & mean$\pm$std & min \\
        \hline
        SVD-Lya    &4.47  &4.45$\pm$0.16 &\textbf{4.20} &21.24$\pm$0.24 &21.02 &\textbf{19.41$\pm$0.11}&19.26 \\
        NS-Lya  &2.88  & 4.51$\pm$0.14 & 4.34 & 21.16$\pm$0.17 & 20.94 &19.65$\pm$0.35  & 19.39\\
        \hline
         MPA-Lya & 2.61 &\textbf{4.39$\pm$0.09} &4.25 & $\textbf{21.11$\pm$0.12}$ &20.95 & \textbf{19.55$\pm$0.20} &\textbf{19.24}\\
         MTP-Lya & \textbf{2.46} & 4.49$\pm$0.13 &4.31 & 21.42$\pm$0.21 &21.24 & 20.55$\pm$0.37&20.12\\
     \hline 
    \end{tabular}
    }
    \label{tab:abla_combination}
\end{table}

\label{sec:abla}

\section{Conclusion}\label{sec:conclusion}

In this paper, we propose two fast methods to compute the differentiable matrix square root and the inverse square root. In the forward pass, the MTP and MPA are applied to approximate the matrix square root, while an iterative Lyapunov solver is proposed to solve the gradient function for back-propagation. A number of numerical tests and computer vision applications demonstrate that our methods can achieve both the fast speed and competitive performances. 


\ifCLASSOPTIONcaptionsoff
  \newpage
\fi



%
\bibliographystyle{IEEEtran}
\bibliography{egbib}
%

%

\begin{IEEEbiography}[{\includegraphics[width=1in,height=1.25in,keepaspectratio]{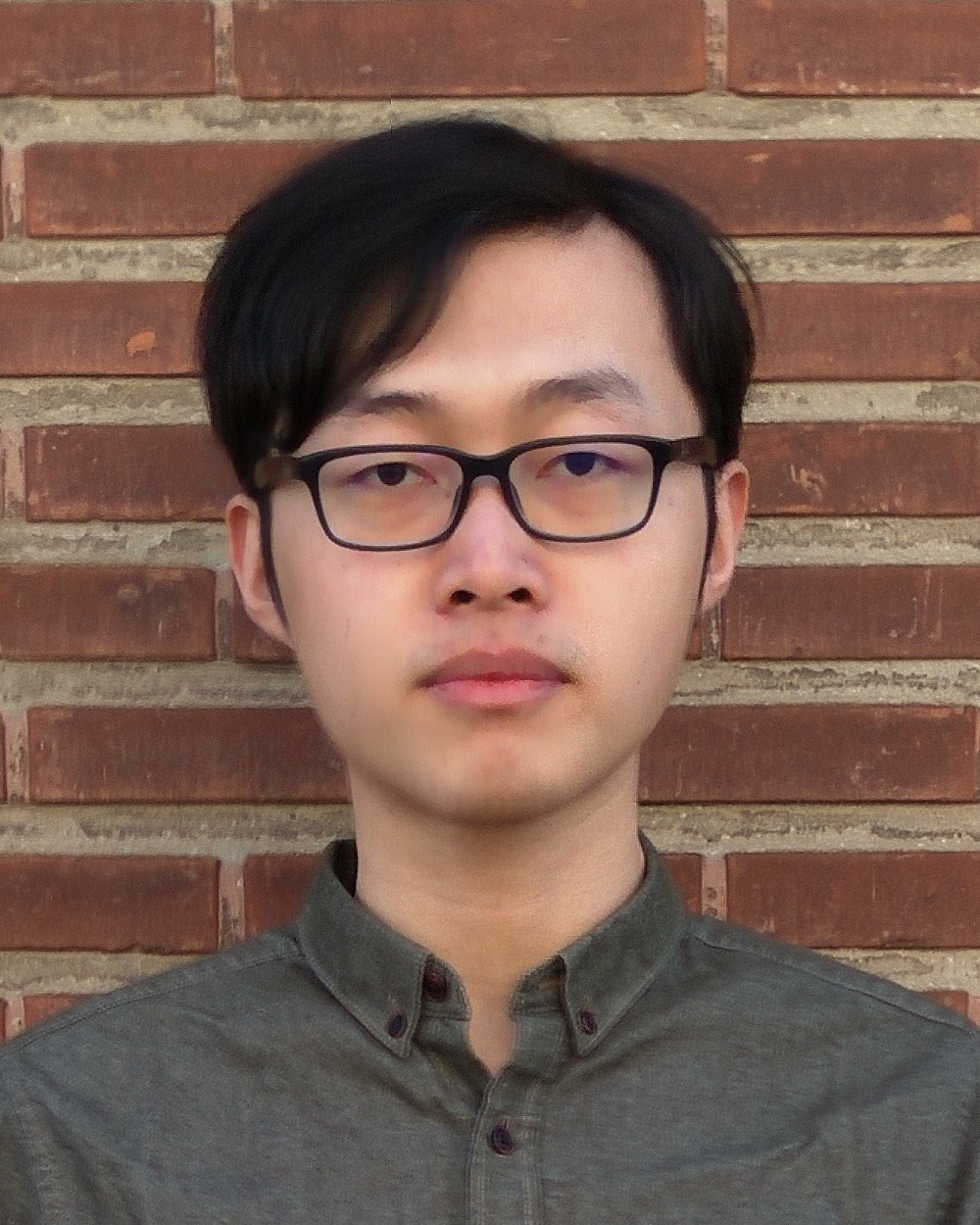}}]{Yue Song}
received the B.Sc. \emph{cum laude} from KU Leuven, Belgium and the joint M.Sc. \emph{summa cum laude} from the University of Trento, Italy and KTH Royal Institute of Technology, Sweden. Currently, he is a Ph.D. student with the Multimedia and Human Understanding Group (MHUG) at the University of Trento, Italy. His research interests are computer vision, deep learning, and numerical analysis and optimization.
\end{IEEEbiography}

\begin{IEEEbiography}[{\includegraphics[width=1in,height=1.25in,keepaspectratio]{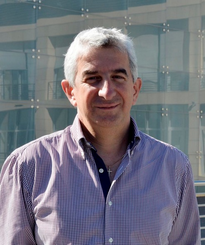}}]{Nicu Sebe} is Professor with the University of
Trento, Italy, leading the research in the areas
of multimedia information retrieval and human
behavior understanding. He was the General
Co- Chair of ACM Multimedia 2013, and the
Program Chair of ACM Multimedia 2007 and
2011, ECCV 2016, ICCV 2017 and ICPR 2020.
He is a fellow of the International Association for
Pattern Recognition.
\end{IEEEbiography}


\begin{IEEEbiography}[{\includegraphics[width=1in,height=1.25in,keepaspectratio]{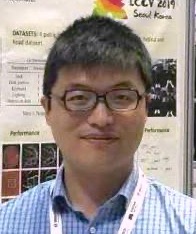}}]{Wei Wang}
is an Assistant Professor of Computer Science at University of Trento, Italy. Previously, after obtaining his PhD from University of
Trento in 2018, he became a Postdoc at EPFL,
Switzerland. His research interests include machine learning and its application to computer
vision and multimedia analysis.
\end{IEEEbiography}

\appendices


\section{Summary of Algorithm}

Algorithm.~\ref{alg:fp} and Algorithm.~\ref{alg:bp} summarize the forward pass (FP) and the backward pass (BP) of our proposed methods, respectively. The hyper-parameter $K$ in Algorithm.~\ref{alg:fp} means the degrees of power series, and $T$ in Algorithm.~\ref{alg:bp} denotes the iteration times.


\begin{algorithm}
\SetAlgoLined
\KwIn{ $\mA$ and $K$}
\KwOut{ $\mA^{\frac{1}{2}}$ or $\mA^{-\frac{1}{2}}$}
 \eIf{MTP}{
  \tcp{FP method is MTP}
  \eIf{Matrix Square Root}{
  $\mA^{\frac{1}{2}}{\leftarrow} \mI {-} \sum_{k=1}^{K} \Big|\dbinom{\frac{1}{2}}{k}\Big| (\mI-\frac{\mA}{||\mA||_{\rm F}})^{k} $\;}
  {
  $\mA^{-\frac{1}{2}}{\leftarrow} \mI {+} \sum_{k=1}^{\infty} \Big|\dbinom{-\frac{1}{2}}{k}\Big| (\mI-\frac{\mA}{||\mA||_{\rm F}})^{k}\;$
  }
 }
 {\tcp{FP method is MPA}
  $M{\leftarrow}\frac{K-1}{2}$, $N{\leftarrow}\frac{K-1}{2}$\; 
  $\mP_{M}{\leftarrow} \mI {-} \sum_{m=1}^{M} p_{m} (\mI-\frac{\mA}{||\mA||_{\rm F}})^{m}$\;
  $\mQ_{N}{\leftarrow} \mI {-} \sum_{n=1}^{N} q_{n} (\mI-\frac{\mA}{||\mA||_{\rm F}})^{n}$\;
  \eIf{Matrix Square Root}{$\mA^{\frac{1}{2}}{\leftarrow}\mQ_{N}^{-1}\mP_{M}$\;}
  {$\mA^{-\frac{1}{2}}{\leftarrow}\mP_{M}^{-1}\mQ_{N}$\;}
 }
 \eIf{Matrix Square Root}{Post-compensate $\mA^{\frac{1}{2}}{\leftarrow}\sqrt{||\mA||_{\rm F}}\cdot\mA^{\frac{1}{2}}$}
 {Post-compensate $\mA^{-\frac{1}{2}}{\leftarrow}\frac{1}{\sqrt{||\mA||_{\rm F}}}\cdot\mA^{-\frac{1}{2}}$}
 \caption{FP of our MTP and MPA for the matrix square root and the inverse square root.}
 \label{alg:fp}
\end{algorithm}

\begin{algorithm}
\SetAlgoLined
\KwIn{$\frac{\partial l}{\partial \mA^{\frac{1}{2}}}$ or $\frac{\partial l}{\partial \mA^{-\frac{1}{2}}}$, 
$\mA^{\frac{1}{2}}$ or $\mA^{-\frac{1}{2}}$, and $T$}
\KwOut{$\frac{\partial l}{\partial \mA}$}
\eIf{Matrix Square Root}
{$\mB_{0}{\leftarrow}\mA^{\frac{1}{2}}$,  $\mC_{0}{\leftarrow}\frac{\partial l}{\partial \mA^{\frac{1}{2}}}$, $i{\leftarrow}0$ \;}
{$\mB_{0}{\leftarrow}\mA^{-\frac{1}{2}}$, $\mC_{0}{\leftarrow}-\mA^{-1}\frac{\partial l}{\partial \mA^{-\frac{1}{2}}}\mA^{-1}$, $i{\leftarrow}0$\;}
Normalize $\mB_{0}{\leftarrow}\frac{\mB_{0}}{||\mB_{0}||_{\rm F}}$, $\mC_{0}{\leftarrow}\frac{\mC_{0}}{||\mB_{0}||_{\rm F}}$\;
 \While{$i<T$}{
  \tcp{Coupled iteration}
  $\mB_{k+1}{\leftarrow}\frac{1}{2} \mB_{k} (3\mI-\mB_{k}^2)$ \;
  $\mC_{k+1}{\leftarrow}\frac{1}{2} \Big(-\mB_{k}^{2}\mC_{k} + \mB_{k}\mC_{k}\mB_{k} + \mC_{k}(3\mI-\mB_{k}^2)\Big)$ \;
  $i{\leftarrow}i+1$\;
 }
 $\frac{\partial l}{\partial \mA}{\leftarrow}\frac{1}{2}\mC_{k}$ \; 
 \caption{BP of our Lyapunov solver for the matrix square root and the inverse square root.}
 \label{alg:bp}
\end{algorithm}

\section{Theoretical Derivation and Proof}

\subsection{Iterative Lyapunov Function Solver}

\begin{dup}[Matrix Sign Function~\cite{higham2008functions}]
  \label{sign_2}
  For a given matrix $\mH$ with no eigenvalues on the imaginary axis, its sign function has the following properties: 1) $sign(\mH)^2=\mI$; 2) if $\mH$ has the Jordan decomposition $\mH{=}\mT\mM\mT^{-1}$, then its sign function satisfies $sign(\mH){=}\mT  sign(\mM) \mT^{-1}$.
\end{dup}
\begin{proof}
  The first property is easy to prove. Consider the SVD of $\mU\mS\mV^{T}=\mH$. As the sign depends on the positiveness of the eigenvale, the square of sign function is computed as:
  \begin{equation}
      sign(\mH)^2= sign(\mS)^2
  \end{equation}
  Since all eigenvalues are real, we have $sign(\mS)^2{=}\mI$, and the first property is proved. The alternative definition of matrix sign function is given by:
  \begin{equation}
      sign(\mH) = \mH(\mH^{2})^{-\frac{1}{2}}
  \end{equation}
  Injecting $sign(\mH){=}\mT  sign(\mM) \mT^{-1}$ into the above equation leads to
  \begin{equation}
  \begin{aligned}
      sign(\mH) &= \mT\mM\mT^{-1}(\mT\mM^2\mT)^{-\frac{1}{2}}\\
                &= \mT\mM\mT^{-1} \mT sign(\mM)\mM^{-1}\mT^{-1}  \\
                &= \mT sign(\mM) \mT^{-1}
  \end{aligned}
  \end{equation}
  The second property gets proved. 
\end{proof}

Now we switch how to derive the iterative solver for matrix sign function in detail. Lemma~\ref{sign_2}.1 shows that $sign(\mH)$ is the matrix square root of the identity matrix. We use the Newton-Schulz iteration to compute $sign(\mH)$ as:
\begin{equation}
    \begin{aligned}
    \mH_{k+1} &{=} = \frac{1}{2}\mH_{k}(3\mI-\mH_{k}^2)\\
    &{=}\frac{1}{2}\begin{bmatrix}
     \mB_{k}{(}3\mI{-}\mB_{k}^2{)} & 3\mC_{k}-\mB_{k}{(}\mB_{k}\mC_{k}{-}\mC_{k}\mB_{k}{)}{-}\mC_{k}\mB_{k}^2 \\
    \mathbf{0} & -\mB_{k}{(}3\mI{-}\mB_{k}^2{)}
    \end{bmatrix}
    \end{aligned} 
\end{equation}
Lemma~\ref{sign_2}.2 indicates an alternative approach to compute the sign function as:
\begin{equation}
\begin{aligned}
    sign(\mH) &=
     sign\Big(\begin{bmatrix}
    \mB & \mC\\
    \mathbf{0} & -\mB
    \end{bmatrix}\Big)\\
    & = \begin{bmatrix}
    \mI & \mX\\
    \mathbf{0} & \mI
    \end{bmatrix} 
    sign\Big(
    \begin{bmatrix}
    \mB & \mathbf{0}\\
    \mathbf{0} & -\mB
    \end{bmatrix} 
    \Big)
    \begin{bmatrix}
    \mI & \mX\\
    \mathbf{0} & \mI
    \end{bmatrix}^{-1} \\
    & = \begin{bmatrix}
    \mI & \mX\\
    \mathbf{0} & \mI
    \end{bmatrix} 
    \begin{bmatrix}
    \mI & \mathbf{0}\\
    \mathbf{0} & -\mI
    \end{bmatrix} 
    \begin{bmatrix}
    \mI & -\mX\\
    \mathbf{0} & \mI
    \end{bmatrix} \\ 
    &=\begin{bmatrix}
    \mI & 2 \mX\\
    \mathbf{0} & -\mI
    \end{bmatrix} 
\end{aligned}
\end{equation}
The above two equations define the coupled iterations and the convergence.

\subsection{Equivalence of two sets of MPA}

\begin{duplicate}
The diagonal MPA $\frac{1}{\sqrt{||\mA||_{\rm F}}}\mS_{N}^{-1}\mR_{M}$ is equivalent to the diagonal MPA $\frac{1}{\sqrt{||\mA||_{\rm F}}}\mP_{M}^{-1}\mQ_{N}$, and the relation $p_{m}{=}-s_{n}$ and $q_{n}{=}-r_{m}$ hold for any $m{=}n$.
\end{duplicate}

\begin{proof}
 Though Pad\'e approximants are derived out of a finite Taylor series, they are asymptotic to their infinite Taylor series~\cite{van2006pade}. Let $f(z){=}(1-z)^{\frac{1}{2}}$ and $f(z)^{-1}{=}(1-z)^{-\frac{1}{2}}$. We have the relation:
 \begin{equation}
     \begin{gathered}
          \frac{1+\sum_{m=1}^{M}r_{m}z^{m}}{1+\sum_{n=1}^{N}s_{n}z^{n}} = f(z)^{-1} +R(z^{M+N+1})\\
          \frac{1-\sum_{m=1}^{M}p_{m}z^{m}}{1-\sum_{n=1}^{N}q_{n}z^{n}} =f(z) +R(z^{M+N+1})\\
     \end{gathered}
 \end{equation}
 where $R(z^{M+N+1})$ is the discarded higher-order term.  Since $f(z)=\frac{1}{f(z)^{-1}}$, we have:
 \begin{equation}
     \frac{1+\sum_{m=1}^{M}r_{m}z^{m}}{1+\sum_{n=1}^{N}s_{n}z^{n}}=\frac{1-\sum_{n=1}^{N}q_{n}z^{n}}{1-\sum_{m=1}^{M}p_{m}z^{m}}.
 \end{equation}
 Now we have two sets of Pad\'e approximants at both sides. Since the numerator and denominator of Pad\'e approximants are relatively prime to each other by definition~\cite{baker1970pade}, the two sets of Pad\'e approximants are equivalent and we have:
 \begin{equation}
     p_{m}=-s_{n},\ q_{n}=-r_{m}
 \end{equation}
 Generalized to the matrix case, this leads to:
 \begin{equation}
     \mP_{M}=\mS_{N},\ \mQ_{N}=\mR_{M}.
 \end{equation}
 Therefore, we also have $\mS_{N}^{-1}\mR_{M}{=}\mP_{M}^{-1}\mQ_{N}$. The two sets of MPA are actually the same representation when $m{=}n$.  
\end{proof}

\subsection{Equivalence of Newton-Schulz Iteration}

\begin{duplicate}
 The one-variable NS iteration of~\cite{huang2019iterative,bini2005algorithms} is equivalent to the two-variable NS iteration of~\cite{li2017second,lin2017improved,higham2008functions}. 
\end{duplicate}
\begin{proof}
  For the two-variable NS iteration, the coupled iteration is computed as:
  \begin{equation}
    \mY_{k+1}=\frac{1}{2}\mY_{k} (3\mI - \mZ_{k}\mY_{k}), \mZ_{k+1}=\frac{1}{2}(3\mI-\mZ_{k}\mY_{k})\mZ_{k}
    \label{prop_ns_two}
  \end{equation}
  where $\mY_{k}$ and $\mZ_{k}$ converge to $\mA^{\frac{1}{2}}$ and $\mA^{-\frac{1}{2}}$, respectively.
  The two variables are initialized as $\mY_{0}{=}\frac{\mA}{||\mA||_{\rm F}}$ and $\mZ_{0}{=}\mI$.

  Since the two variables have the relation $\mZ_{k}^{-1}\mY_{k}{=}\frac{\mA}{||\mA||_{\rm F}}$, we can replace $\mY_{k}$ in~\cref{prop_ns_two} with $\mZ_{k}\frac{\mA}{||\mA||_{\rm F}}$:
  \begin{equation}
      \mZ_{k+1}=\frac{1}{2}(3\mI-\mZ_{k}^{2}\frac{\mA}{||\mA||_{\rm F}})\mZ_{k}
  \end{equation}
  Notice that $\mA$ and $\mZ_{k}$ have the same eigenspace and their matrix product commutes, \emph{i.e.,} $\mA\mZ_{k}{=}\mZ_{k}\mA$. Therefore, the above equation can be further simplified as:
  \begin{equation}
      \mZ_{k+1}=\frac{1}{2}(3\mZ_{k}-\mZ_{k}^{3}\frac{\mA}{||\mA||_{\rm F}})
  \end{equation}
  As indicated above, the two seemingly different NS iterations are in essence equivalent.
\end{proof}

\section{Baselines}
\label{app:baselines}

In the experiment section, we compare our proposed two methods with the following baselines:
\begin{itemize}
    \item Power Iteration (PI). It is suggested in the original So-ViT to compute only the dominant eigenpair. 
    \item SVD-PI~\cite{wang2019backpropagation} that uses PI to compute the gradients of SVD.
    \item SVD-Taylor~\cite{wang2021robust,song2021approximate} that applies the Taylor polynomial to approximate the gradients.
    \item SVD-Pad\'e~\cite{song2021approximate} that proposes to closely approximate the SVD gradients using Pad\'e approximants. Notice that our MTP/MPA used in the FP is fundamentally different from the Taylor polynomial or Pad\'e approximants used in the BP of SVD-Pad\'e. For our method, we use Matrix Taylor Polynomial (MTP) and Matrix Pad\'e Approximants (MPA) to derive the matrix square root in the FP. For the SVD-Pad\'e, they use scalar Taylor polynomial and scalar Pad\'e approximants to approximate the gradient $\frac{1}{\lambda_{i}-\lambda_{j}}$ in the BP. That is to say, their aim is to use the technique to compute the gradient and this will not involve the back-propagation of Taylor polynomial or Pad\'e approximants. 
    \item NS iteration~\cite{schulz1933iterative,higham2008functions} that uses the Newton-Schulz iteration to compute the matrix square root. It has been widely applied in different tasks, including covariance pooling~\cite{li2018towards} and ZCA whitening~\cite{huang2018decorrelated}. We note that although~\cite{huang2019iterative} and~\cite{higham2008functions} use different forms of NS iteration, the two representations are equivalent to each other (see the proof in the paper). The modified NS iteration in~\cite{huang2019iterative} just replaces $\mY_{k}$ with $\mZ_{k}\mA$ and re-formulates the iteration using one variable. The computation complexity is still the same.
\end{itemize}

As the ordinary differentiable SVD suffers from the gradient explosion issue and easily causes the program to fail, we do not take it into account for comparison.

Unlike previous methods such as SVD and NS iteration, our MPA-Lya/MTP-Lya does not have a consistent FP and BP algorithm. However, we do not think it will bring any caveat to the stability or performance. Our MTP and MPA do not need coupled iteration in the FP and always have gradient back-propagating through $\mathbf{A}^{\frac{1}{2}}$ or $\mathbf{A}^{-\frac{1}{2}}$ in the BP, which could guarantee the training stability. Moreover, our ablation study implies that our BP Lyapunov solver approximates the real gradient very well (\emph{i.e.,} $||\mB_{k}{-}\mI||_{\rm F}{<}3e{-}7$ and $||0.5\mC_{k}{-}\mX||_{\rm F}{<}7e{-}6$). Also, our extensive experiments demonstrate the superior performances. In light of these experimental results, we argue that as long as the BP algorithm is accurate enough, the inconsistency between the BP and FP is not an issue.


\section{Experimental Settings}
\label{app:exp_set}

All the source codes are implemented in Pytorch. For the SVD methods, the forward eigendecomposition is performed on the CPU using the official Pytorch function \textsc{torch.svd}, which calls the LAPACK's routine \textit{gesdd} that uses the Divide-and-Conquer algorithm for the fast calculation. All the numerical tests are conducted on a single workstation equipped with a Tesla K40 GPU and a 6-core Intel(R) Xeon(R) GPU @ 2.20GHz.

For our method throughout all the experiments, in the forward pass, we match the MTP to the power series of degree $11$ and set the degree for both numerator and denominator of our MPA as $5$. We keep iterating $8$ times for our backward Lyapunov solver.

Now we turn to the implementation details for each experiment in the paper. 

\subsection{Decorrelated Batch Normalization}

\begin{figure}[htbp]
    \centering
    \includegraphics[width=0.5\linewidth]{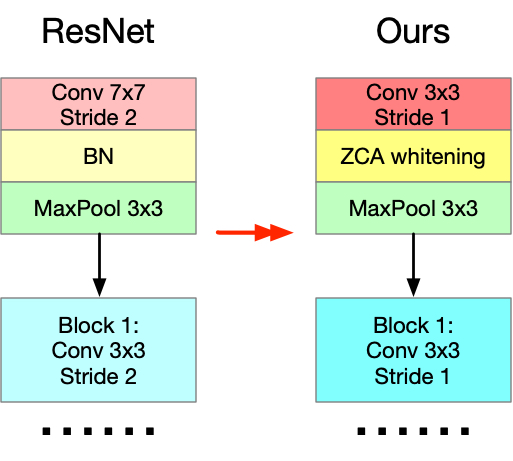}
    \caption{The architecture changes of ResNet models in the experiment of ZCA whitening. The decorrelated batch normalization layer is inserted after the first convolutional layer. The kernel sizes, the stride of the first convolution layer, and the stride of the first ResNet block are changed correspondingly.}
    \label{fig:arch_BN}
\end{figure}

Fig.~\ref{fig:arch_BN} displays the detailed architecture changes of ResNet. Suggested by~\cite{wang2020deep}, we truncate the Taylor polynomial to degree $20$ for SVD-Taylor. To make Pad\'e approximant match the same degree with Taylor polynomial, we set the degree of both numerator and denominator to $10$ for SVD-Pad\'e. For SVD-PI, the iteration times are also set as $20$. For the NS iteration, according to the setting in~\cite{li2018towards,huang2018decorrelated}, we set the iteration times to $5$. The other experimental settings follow the implementation in~\cite{wang2021robust}. We use the workstation equipped with a Tesla K40 GPU and a 6-core Intel(R) Xeon(R) GPU @ 2.20GHz for training. Notice that in our previous conference paper, we first calculate the matrix square root $\mA^{\frac{1}{2}}$ and then compute $\mX_{whitend}$ by solving the linear system $\mA^{\frac{1}{2}}\mX_{whitend}{=}\mX$. Thanks to the algorithm extension to the inverse square root, we can directly computes $\mA^{-\frac{1}{2}}$ in this paper.


\subsection{Second-order Vision Transformer}

We use 8 Tesla G40 GPUs for distributed training and the NVIDIA Apex mixed-precision trainer is used. Except that the spectral layer uses the single-precision (\emph{i.e.,} float32), other layers use the half-precision (\emph{i.e.,} float16) to accelerate the training. Other implementation details follow the experimental setting of the original So-ViT~\cite{xie2021so}. Following the experiment of covariance pooling for CNNs~\cite{song2021approximate}, the degrees of Taylor polynomial are truncated to $100$ for SVD-Taylor, and the degree of both the numerator and denominator of Pad\'e approximants are set to $50$ for SVD-Pad\'e. The iteration times of SVD-PI are set to $100$. In the experiment of covariance pooling, more terms of the Taylor series are used because the covariance pooling meta-layer requires more accurate gradient estimation~\cite{song2021approximate}.  

For the SVD-based methods, usually the double-precision is required to ensure an effective numerical representation of the eigenvalues. Using a lower precision would make the model fail to converge at the beginning of the training~\cite{song2021approximate}. This is particularly severe for vision transformers which are known slow and hard to converge in the early training stage. One may consider to cast the tensor into double-precision (64 bits) to alleviate this issue. However, this will trigger much larger gradient and introduce round-off errors when the gradient is passed to previous layer in half-precision (16 bits). To avoid this caveat, we first apply the NS iteration to train the network for $50$ epochs, then switch to the corresponding SVD method and continue the training till the end. This hybrid approach can avoid the non-convergence of the SVD methods at the beginning of the training phase. 


\subsection{Global Covariance Pooling}


For the experiment on large-scale and fine-grained image recognition, we refer to~\cite{song2021approximate} for all the experimental settings. In the video action recognition experiment~\cite{gao2021temporal}, the iteration time for NS iteration is set as $5$. Othe implementation details are unchanged. 






\subsection{Neural Style Transfer}

\begin{table*}[htbp]
\caption{The detailed LPIPS~\cite{zhang2018perceptual} score and user preference (\%) on each subset of Artworks dataset.}
    \centering
    \resizebox{0.9\linewidth}{!}{
    \begin{tabular}{r|c|c|c|c|c|c|c|c|c|c}
    \hline
    \multirow{2}*{Methods} & \multicolumn{5}{c|}{LPIPS~\cite{zhang2018perceptual} Score ($\uparrow$)} & \multicolumn{5}{c}{User Preference ($\uparrow$)} \\
    \cline{2-11}
          & Cezanne & Monet & Vangogh & Ukiyoe & Average & Cezanne & Monet & Vangogh & Ukiyoe & Average  \\
        \hline
        SVD-Taylor &0.4937 &0.4820 &\textbf{0.6074} &0.5274 & 0.5276 & 15 &16 &\textbf{25} &9 &16.25\\
        SVD-Pad\'e &0.6179 &0.4783 &0.5307 &0.5419 &0.5422  & \textbf{28} &13 &15 & 21&  19.25\\
        NS iteration &0.5328 &\textbf{0.5329} &0.5386 &0.6270 & 0.5578&11 & 18 &21 & 18 & 17.00\\
        \hline
        Our MPA-Lya &\textbf{0.6332} &0.5291 &0.4511 &\textbf{0.6325} & \textbf{0.5615} & 25 &\textbf{29} &18 &\textbf{27} & \textbf{24.75}\\
        Our MTP-Lya &0.6080 &0.4826 &0.4796 &{0.6253} & 0.5489 & 17&21 &17 & 19 & 18.50 \\
    \hline
    \end{tabular}
    }
    \label{tab:style_transfer_full}
\end{table*}

\begin{figure*}[htbp]
    \centering
    \includegraphics[width=0.85\linewidth]{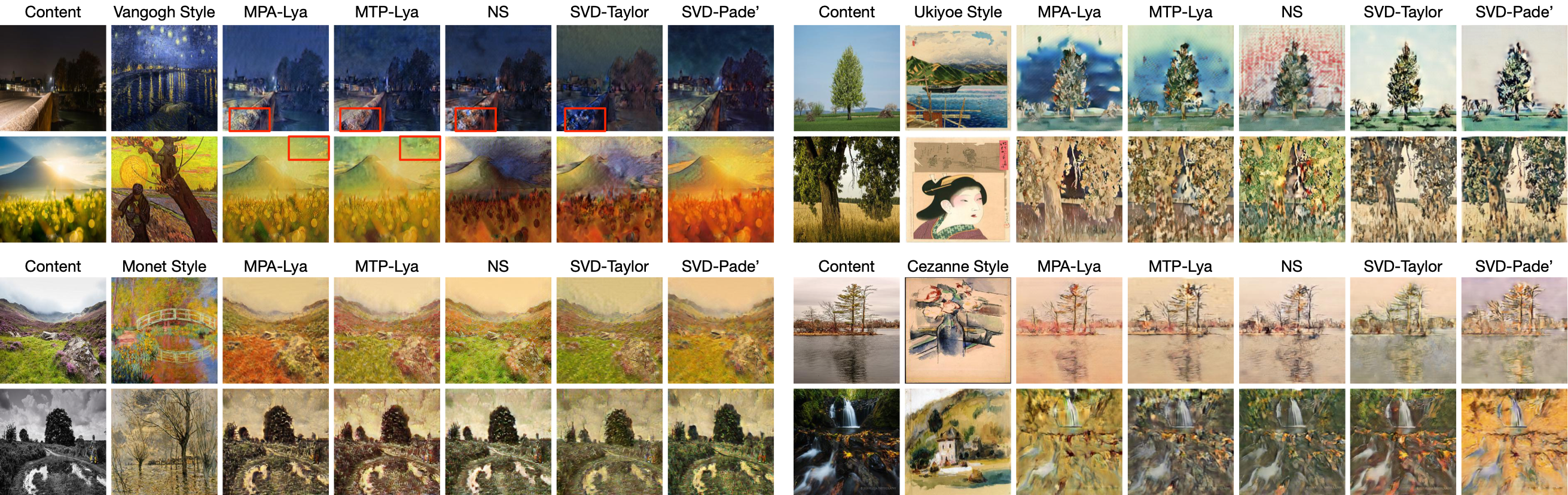}
    \caption{More exemplary visualizations on Artworks~\cite{isola2017image} dataset. Our methods generate sharper images with more coherent style and better visual appeal. The red rectangular indicates regions with subtle details.}
    \label{fig:style_transfer_visual_large}
\end{figure*}


For the loss functions, we follow the settings in~\cite{cho2019image} and use the cycle-consistent reconstruction loss in both the latent and the pixel space. The image is resized to the resolution of $216{\times}216$ before passing to the network, and the model is trained for $100,000$ iterations. The batch size is set to $4$.

Table~\ref{tab:style_transfer_full} and Fig.~\ref{fig:style_transfer_visual_large} present the detailed quantitative evaluation and more visual comparison, respectively. As suggested in~\cite{li2017universal,wang2020diversified}, we use the LPIPS~\cite{zhang2018perceptual} score and the user preference as the evaluation metrics. For the LPIPS metric, we compute the score between each pair of transferred image and the content image. A higher LPIPS score implies that the image carries less content information but more style information. For the user study, we randomly select $100$ images from each dataset and ask $20$ volunteers to vote for the image that characterizes more the style information. In some cases where the volunteer thinks none of the images correctly carries the style, he/she can abstain and does not vote for any one.

\section{Comparison of Lyapunov Solver against Implicit Function and Automatic Differentiation}

Besides our proposed custom Lyapunov gradient solver, one may consider alternative gradient computation schemes, such as reverse-mode automatic differentiation (RMAD) and implicit function (IF). For the RMAD, the backward pass indeed takes roughly the same operation costs as the forward pass. Considering that our MPA uses two sets of matrix power polynomials and one matrix inverse, using RMAD for the gradient computation would be less efficient than the Lyapunov solver which only involves matrix multiplications. Moreover, the gradient of some intermediate variables of MPA would be calculated in the RMAD, which would further increase unnecessary memory costs. For the IF, the function for matrix square root can be defined as $f(\mA,\mA^\frac{1}{2})=(\mA^{\frac{1}{2}})^2-\mA$ where $\mA^{\frac{1}{2}}$ can be regarded as a function of $\mA$. Performing implicit differentiation and multiplying both sides with $\frac{\partial l}{\partial \mA^{\frac{1}{2}}}$ would lead to the gradient equation $\frac{\partial l}{\partial \mA}=-(\frac{\partial f}{\partial \mA^{\frac{1}{2}}})^{-1}\frac{\partial f}{\partial \mA}\frac{\partial l}{\partial \mA^{\frac{1}{2}}}$. The memory usage of IF should be small since only the gradient of $f$ is introduced in the computation. However, the time cost can be high due to the function gradient evaluation $\frac{\partial f}{\partial \mA}$ and $\frac{\partial f}{\partial \mA^{\frac{1}{2}}}$ as well as the matrix inverse computation.

\begin{table}[htbp]
    \centering
    \caption{Backward time and speed comparison for batched matrices of size $64{\times}64{\times}64$. We use MPA for forward pass, and the evaluation is averaged on $1,000$ randomly generated matrices.}
    \begin{tabular}{c|c|c}
    \toprule
    Method & Speed (ms) & Memory (MB) \\
     \hline
         Lyapunov   &\textbf{2.19} & \textbf{1.99}\\
         RMAD &5.69 & 3.08  \\
         IF    &4.71 & 2.03\\ 
    \bottomrule
    \end{tabular}
    \label{tab:rmad_if}
\end{table}

Table~\ref{tab:rmad_if} compares the speed and memory consumption. Our Lyapunov solver outperforms both schemes in terms of speed and memory. The memory usage of IF is competitive, which also meets our expectation. In general, our Lyapunov-based solver can be viewed as a well-optimized RMAD compiler with the least memory and time consumption.

\section{Stability of Pad\'e Approximants}
\label{app:stability_pade}

When there is the presence of spurious poles~\cite{stahl1998spurious,baker2000defects}, the Pad\'e approximants are very likely to suffer from the well-known defects of instability. The spurious poles mean that when the approximated function has very close poles and zeros, the corresponding Pad\'e approximants will also have close poles and zeros. Consequently, the Pad\'e approximants will become very unstable in the region of defects (\emph{i.e.,} when the input is in the neighborhood of poles and zeros). Generalized to the matrix case, the spurious poles can happen when the determinant of the matrix denominator is zero (\emph{i.e.} $\det{(\mQ_{N})}=0$). 

However, in our case, the approximated function for matrix square root is $(1-z)^{\frac{1}{2}}$ for $|z|<1$, which only has one zero at $z=1$ and does not have any poles. For the inverse square root, the approximated function $(1-z)^{-\frac{1}{2}}$ has one pole but does not have an zeros. Therefore, the spurious pole does not exist in our approximation and there are no defects of our Pad\'e approximants.

Now we briefly prove this claim for the matrix square root. The proof for the inverse square root can be given similarly, and we omit it here for conciseness. Consider the denominator of our Pad\'e approximants:
\begin{equation}
    \mQ_{N}=  \mI - \sum_{n=1}^{N} q_{n} (\mI-\frac{\mA}{||\mA||_{\rm F}})^{n}
    \label{eq:QN_deno}
\end{equation}
Its determinant is calculated as:
\begin{equation}
    \det{(\mQ_{N})}=\prod_{i=1}(1-\sum_{n=1}^{N} q_{n}(1-\frac{\lambda_{i}}{\sqrt{\sum_{i}\lambda_{i}^{2}}})^{n})
    \label{eq:QN_det}
\end{equation}
The coefficients $q_{n}$ of our $[5,5]$ Pad\'e approximant are pre-computed as $[2.25,-1.75,0.54675,-0.05859375,0.0009765625]$. Let $x_{i}$ denotes $(1-\frac{\lambda_{i}}{\sqrt{\sum_{i}\lambda_{i}^{2}}})$. Then $x_{i}$ is in the range of $[0,1]$, and we have:
\begin{equation}
\begin{gathered}
    f(x_{i})=1-2.25x_{i}+1.75x^2_{i}-0.54675x^3_{i}+\\+0.05859375x^{4}_{i}-0.0009765625x^{5}_{i}; \\ 
    \det{(\mQ_{N})}=\prod_{i=1}(f(x_{i})).
\end{gathered}
\label{eq:QN_nonzero}
\end{equation}
The polynomial $f(x_{i})$ does not have any zero in the range of $x{\in}[0,1]$. The minimal is $0.0108672$ when $x=1$. This implies that $\det{(\mQ_{N})}\neq0$ always holds for any $\mQ_{N}$ and our Pad\'e approximants do not have any pole. Accordingly, there will be no spurious poles and defects. Hence, our MPA is deemed stable. Throughout our experiments, we do not encounter any instability issue of our MPA.

\end{document}